\documentclass{article}

\usepackage{microtype}
\usepackage{graphicx}
\usepackage{booktabs} 

\usepackage{hyperref}

\usepackage{subcaption}
\usepackage{comment}
\usepackage{enumitem}

\usepackage{amsmath,amsfonts,bm}

\newcommand{\newterm}[1]{{\emph{#1}}}

\def\1{\bm{1}}

\def\rvx{{\mathbf{x}}}

\def\rvz{{\mathbf{z}}}

\def\vu{{u}}
\def\vv{{v}}

\def\vx{{x}}
\def\vz{{z}}

\DeclareMathAlphabet{\mathsfit}{\encodingdefault}{\sfdefault}{m}{sl}
\SetMathAlphabet{\mathsfit}{bold}{\encodingdefault}{\sfdefault}{bx}{n}

\newcommand{\pdata}{p_{\rm{data}}}

\newcommand{\E}{\mathbb{E}}

\DeclareMathOperator*{\argmin}{arg\,min}

\renewcommand{\E}{\mathbb{E}}

\usepackage{xcolor}

\usepackage{amsmath}
\usepackage{amssymb}
\usepackage{amsthm}
\newtheorem{defi}{Definition}[section]
\newtheorem{theo}[defi]{Theorem}
\newtheorem{exa}[defi]{Example}

\usepackage[accepted]{innf2021}

\icmltitlerunning{Copula-Based Normalizing Flows}

\begin{document}

\twocolumn[
\icmltitle{              Copula-Based Normalizing Flows}



\icmlsetsymbol{equal}{*}

\begin{icmlauthorlist}
\icmlauthor{Mike Laszkiewicz}{rub}
\icmlauthor{Johannes Lederer}{rub}
\icmlauthor{Asja Fischer}{rub}
\end{icmlauthorlist}

\icmlaffiliation{rub}{Department of Mathematics, Ruhr University, Bochum, Germany}

\icmlcorrespondingauthor{Mike Laszkiewicz}{Mike.Laszkiewicz@rub.de}

\icmlkeywords{Machine Learning, ICML}

\vskip 0.3in
]



\printAffiliationsAndNotice{\icmlEqualContribution} 

\begin{abstract}
Normalizing flows, which learn a distribution by transforming the data to samples from a Gaussian base distribution, have proven powerful density approximators.
But their expressive power is limited by this choice of the base distribution.
We, therefore, propose to generalize the base distribution to a more elaborate copula distribution
to capture the properties of the target distribution more accurately.
In a first empirical analysis, we demonstrate that this replacement 
can dramatically improve the vanilla normalizing flows in terms of flexibility, stability, and effectivity for heavy-tailed data.
Our results suggest that the improvements are related to an increased  local Lipschitz-stability of the learned flow.
\end{abstract}

\section{Introduction}
\label{sec:intro}
\newterm{Normalizing Flows} (NFs) are a recently developed class of density estimators, which aim to transform the distribution of interest $P_\rvx$  to some tractable base distribution $P_\rvz$. 
Using the change of variables formula, this allows for exact likelihood computation, which is in contrast to other deep generative models such as Variational Autoencoders \citep{kingma2013auto} or Generative Adversarial Networks \citep{goodfellow2014generative}. Impressive estimation results, especially in the field of natural image generation, have lead to great popularity of these deep generative models. Motivated by this success, much effort has been put into the development of new parametric classes of NFs to make them even more performant \cite{dinh2014nice, dinh2016density,chen2019residual,papamakarios2017masked,grathwohl2018ffjord}. 
However, our theoretical understanding has not developed at the same speed, which, so we claim, slows down further progress in the development of powerful NF architectures.

Fortunately, very recent works have addressed theoretical limitations of these methods: One important limitation is the expressive power of NFs.
Because they are based on the change of variables formula, 
the learned transformations are required to be diffeomorphisms. As a consequence, a NF with bounded Lipschitz constant is unable to map one distribution $P_\rvx$ to a lighter-tailed distribution $P_\rvz$ \citep{wiese2019copula, pmlr-v119-jaini20a}. 
Therefore, since vanilla NFs are implemented using an isotropic Gaussian base distribution, they are unable to learn heavy-tailed distributions, which are known to appear in natural image data \citep{zhu-longtails, vanhorn2017devil, range_loss}. 
This is in conflict with observations recently made by \citet{pmlr-v130-behrmann21a}: Even though NFs are typically designed to obey invertibility, this property is often violated in practice. 
This is due to numerical inaccuracies, which are promoted by a large Bi-Lipschitz constant. Bounding the Bi-Lipschitz constant, however, conflicts with the previously mentioned theoretical requirements needed to avoid a limited expressiveness of the NF. 

These findings emphasize the high importance of choosing an appropriate base distribution for NFs. We therefore propose to generalize the isotropic Gaussian to a much broader class of distributions---the class of \newterm{copula} distributions. Copulae are a well-known concept in statistical theory and are being used to model complex distributions in finance, actuarial science, and extreme-value theory \citep{genest, joe2014dependence, elementsofcopula}. Broadly speaking, a copula is some function that couples marginal distributions to a multivariate joint distribution. Hence, copulae allow for flexible multivariate modeling with marginals stemming from a huge range of suitable classes. 
This allows for example to formulate NF base distributions that combine heavy-tailed marginals---as proposed by \citet{wiese2019copula, pmlr-v119-jaini20a,alexanderson2020robust}---with light-tailed marginals. 
This paper presents a novel  framework for choosing the base distribution of NFs building on the well-studied theory of copulae. A first empirical investigation demonstrates the benefits brought by this approach.
Our experimental analysis on toy data reveals that using even the most simple copula model---the Independence Copula---we are able to outperform the vanilla NF approach, which uses an isotropic Gaussian base distribution.
The resulting NF converges faster, is more robust, and achieves an overall better test performance. In addition, we show that the learned transformation has a better-behaved functional form in the sense of a more stable local Lipschitz continuity.
\section{Background}
\label{sec:background}
In this section, we quickly review some background knowledge about NFs (Section~\ref{sec:NFs}), followed by an introduction to copula theory (Section~\ref{sec:copula}).
\subsection{Normalizing Flows}
\label{sec:NFs} 
Density estimation via NFs revolve around learning a diffeomorphic transformation $T_\theta$ that maps some unknown target distribution $P_\rvx$ to a known and tractable base distribution $P_\rvz$. At the cornerstone of NFs is the \newterm{change of variables formula}
\begin{equation}
        p_\theta(\vx) = p_\rvz\bigl(T_\theta (\vx)\bigr) \bigl\vert \det J_{T_\theta}(\vx) \bigr\vert\quad \text{for }\vx\in\mathbb{R}^D\enspace , \label{eq:changeofvar}
\end{equation}
which relates the evaluation of the estimated density $p_\theta$ of $\rvx\sim P_\rvx$ to the evaluation of the base density $p_\rvz$, of $T_\theta(\vx)$, and of $\det J_{T_\theta}(\vx)$. 
By composing simple diffeomorphic building blocks $T_\theta:=T_{\theta,l} \circ \cdots \circ T_{\theta, 1}$, we are able to obtain expressive transformations, while presuming diffeomorphy and computational tractablity of the building blocks. Due to the tractable PDF in \eqref{eq:changeofvar}, we are able to train the model via maximum likelihood estimation (MLE)
\begin{equation}
        \label{eq:mletraining}
        \hat{\theta} \in \argmin_{\theta} \E_{\pdata} \bigl[ -\log p_\theta (\rvx) \bigr]\enspace ,
    \end{equation}
where $p_{\operatorname{data}}$ is the PDF of the empirical distribution of $\rvx$. A comprehensive overview of NFs, including the exact parameterizations of certain flow models $T_\theta$, computational aspects, and more, can be found in \citet{kobyzev2020normalizing} and \citet{papamakarios2019normalizing}.

\subsection{Copulae}
\label{sec:copula}
A completely different approach of density estimation, which has mostly been left unrelated to NFs, is the idea of copulae. 
\begin{defi}[Copula] \label{defi:copula}
A copula is a multivariate distribution with CDF $C:[0,1]^D \rightarrow [0, 1]$ that has standard uniform marginals, i.e. the marginals $C_j$ of $C$ satisfy $C_j \sim U[0,1]$.
\end{defi}
The fundamental idea behind copula theory is that we can associate every distribution with a uniquely defined copula $C$. Vice versa, given $D$ marginal distributions, each copula $C$ defines a multivariate distribution with the given marginals. Formally, this is known as \newterm{Sklar's Theorem} \citep{sklar, sklarproof}.
\begin{theo}[Sklar's Theorem] Taken from \citet{elementsofcopula}. \label{sklarstheorem}
\begin{enumerate}[nolistsep]
    \item For any $D$-dimensional CDF $F_\rvz$ with marginal CDFs $F_{\rvz_1},\dots, F_{\rvz_D}$, there exists a copula $C$ such that
    \begin{equation}
    \label{eq:sklar_df}
        F_\rvz (\vz) = C\bigl( F_{\rvz_1}(\vz_1), \dots, F_{\rvz_D}(\vz_D) \bigr) \enspace 
    \end{equation}
    for all $\vz\in\mathbb{R}^D$. The copula is uniquely defined on $\mathcal{U}:=\prod_{j=1}^D \operatorname{Im}(F_{\rvz_j})$, where $\operatorname{Im}(F_{\rvz_j})$ is the image of $F_{\rvz_j}$. For all $\vu\in\mathcal{U}$ it is given by 
    \begin{equation}
        \label{eq:sklar_copula}
        C(\vu) = F_\rvz \bigl( F_{\rvz_1}^{\leftarrow}(\vu_1), \dots , F_{\rvz_D}^{\leftarrow}(\vu_D) \bigr)\enspace ,
    \end{equation}
    where $F_{\rvz_j}^{\leftarrow}$ are the right-inverses of $F_{\rvz_j}$.    
    \item Conversely, given any $D$-dimensional copula $C$ and marginal CDFs $F_{\rvz_1}, \dots F_{\rvz_D}$, a function $F_{\rvz}$ as defined in \eqref{eq:sklar_df} is a $D$-dimensional CDF with marginals $F_{\rvz_1},\dots ,F_{\rvz_D}$.
\end{enumerate}
\end{theo}
Part~1 of Sklar's Theorem finds much application in statistical dependency analysis \citep{joe2014dependence}. In contrast to classical dependency measures, such as Pearson correlation, copulae are a more flexible tool that allow the decoupling of the marginals and the dependency structure. Part~2 of Sklar's Theorem is of relevance for statistical modeling, and more precisely, to define multivariate distributions. Given marginal distributions, which are typically much easier to estimate than the full joint distribution, and a copula $C$ we can ``couple'' the marginals and the dependency structure to a multivariate joint distribution. 
This perspective finds various applications in the context of finance and related disciplines that need to take heavy tails and tail dependencies into account, see \citet{genest} for an overview.
In Section~\ref{sec:examples_copula} of the Appendix we give some illustrative examples and further details about properties of copula distributions. 

By differentiating~\eqref{eq:sklar_df}, we obtain the PDF of $\rvz$ as 
\begin{equation}
    \label{eq:copuladensity}
    p_{\rvz}(\vz) = c\bigl( F_{\rvz_1}(\vz_1), \dots , F_{\rvz_D}(\vz_D) \bigr) \prod_{j=1}^D p_{\rvz_j}(\vz_j)\enspace, 
\end{equation}
where $c,\, p_{\rvz_1},\dots,\, p_{\rvz_D}$ are the PDFs of $C, \, F_{\rvz_1},\dots ,\, F_{\rvz_D}$, respectively.

\section{NFs With Copula-Base Distributions}
In this paper, we propose to
employ copulae to model a flexible, yet appropriate base distribution with the goal of gaining a NF that is able to solve the limitations of NFs discussed in Section~\ref{sec:intro}.
We expect to gain powerful and robust PDF approximators by combining different marginals and properties of theoretical sound copulae (see for instance Chapter~8 in \citet{joe2014dependence}) with NFs, which allow for the
estimation of complex densities.

\subsection{A General Framework} \label{sec:framework}
We propose to replace
the isotropic Gaussian base distribution in the vanilla NF framework by a more flexible copula distribution. 
Importantly, we want to learn a base distribution that is able to represent the tail behavior of the distribution of $\rvx$.
For training a NF with a copula base distribution we build on the fact that we can write the PDF of the latent variables as written in \eqref{eq:copuladensity}.
This requires two estimation steps:
First, we need to estimate the marginal distributions $F_{\rvz_1}, \dots ,F_{\rvz_D}$, which can further be used to calculate  the marginal densities $p_{\rvz_1}, \dots ,p_{\rvz_D}$. Secondly, we need to estimate the copula density $c$. 
A popular approach for estimating the density in \eqref{eq:copuladensity} is to employ the method of inference functions for margins (IFM) \citep{Joe_Xu_1996}, which sequentially estimates the marginals using MLE first, and then employs these marginals to estimate the copula using MLE.

It is important to note that in contrast to standard applications of copula theory, we do not aim at
estimating the full data generating distribution based on \eqref{eq:copuladensity}. Instead, following the investigations by \citet{pmlr-v119-jaini20a}, our goal is to capture the tail-behavior of $\rvx$. Hence, we propose to learn surrogate marginals $\rvz_1, \dots, \rvz_D$ that are able to represent the tailedness of the marginals $\rvx_1, \dots ,\rvx_D$. By combining these marginals with some simple copula structure, such as the Gaussian Copula or the \newterm{Independence Copula} (see \eqref{eq:prod_dist} below), we are able to create a joint distribution that represents the marginal tail behavior of $\rvx$. 

The proposed adjustment can be applied to any existing NF architecture as long as \eqref{eq:copuladensity} remains tractable. However, as the main goal of the base distribution is not to fully estimate the target but to represent the tail behavior of $\rvx$, we can restrict ourselves to tractable parametric marginal distributions and copulae. 

\subsection{Experimental Analysis}
\label{sec:experiments}
In this section, we investigate the
benefits of the proposed approach by analyzing a toy problem. In the following experiments, we employ the
framework proposed in Section~\ref{sec:framework} using the most simple copula: the Independence Copula, i.e.~we consider a base distribution $\rvz$ with PDF 
\begin{equation}
    p_{\rvz}(\vz)= \prod_{j=1}^D p_{\rvz_j}(\vz_j), \quad \vz \in \prod_{j=1}^D \operatorname{supp}(\rvz_j)\enspace . \label{eq:prod_dist}
\end{equation}
Note that by plugging Gaussian marginals in \eqref{eq:prod_dist}, we would obtain the vanilla NF.
We consider a training set generated from a 2-dimensional heavy-tailed distribution,\footnote{all computational details can be found in Section~\ref{sec:comp_details} of the Appendix} which has standardized t-distributed marginals $\rvx_1, \rvx_2\sim t_2(0, 1)$ with $2$ degrees of freedom. The corresponding copula is a \newterm{Gumbel Copula} with parameter $\rho=2.5$, i.e. 
\begin{equation*}
    C(\vu):= \exp\bigl( 
        -((-\log(\vu_1))^\rho + (-log(\vu_2))^\rho)^{1/\rho}
    \bigr) \enspace .
\end{equation*}

As a proof of concept we compare the estimation of this heavy-tailed distribution using a NF\footnote{We are using a 3-layered MAF \cite{papamakarios2017masked}. Further computational details can be found in Section~\ref{sec:comp_details} of the Appendix.} with an isotropic Gaussian base distribution, and with 3 different heavy-tailed base distributions constructed via \eqref{eq:prod_dist}. We consider the following heavy-tailed marginals:
\begin{enumerate}[nolistsep]
    \item 
    Laplace$(0, 4)$ and $t_5(0, 2)$. We call this case \emph{heavierTails} because one marginal is heavy-tailed; 
    \item 
    $t_5(0, 1)$ and $t_5(0, 1)$. We call this case \emph{correctFamily} since both marginals stem from the same parametric class as the exact marginals;
    \item $t_2(0,1)$ and $t_2(0,1)$. We call this case \emph{exactMarginals}.
\end{enumerate}
Samples from the target distribution and from the different base distributions are visualized in Figure~\ref{fig:base_dist} and~\ref{fig:target_dist} in the Appendix. 

\begin{figure}
    \centering
    \begin{subfigure}{0.23\textwidth}
       \includegraphics[width=\textwidth]{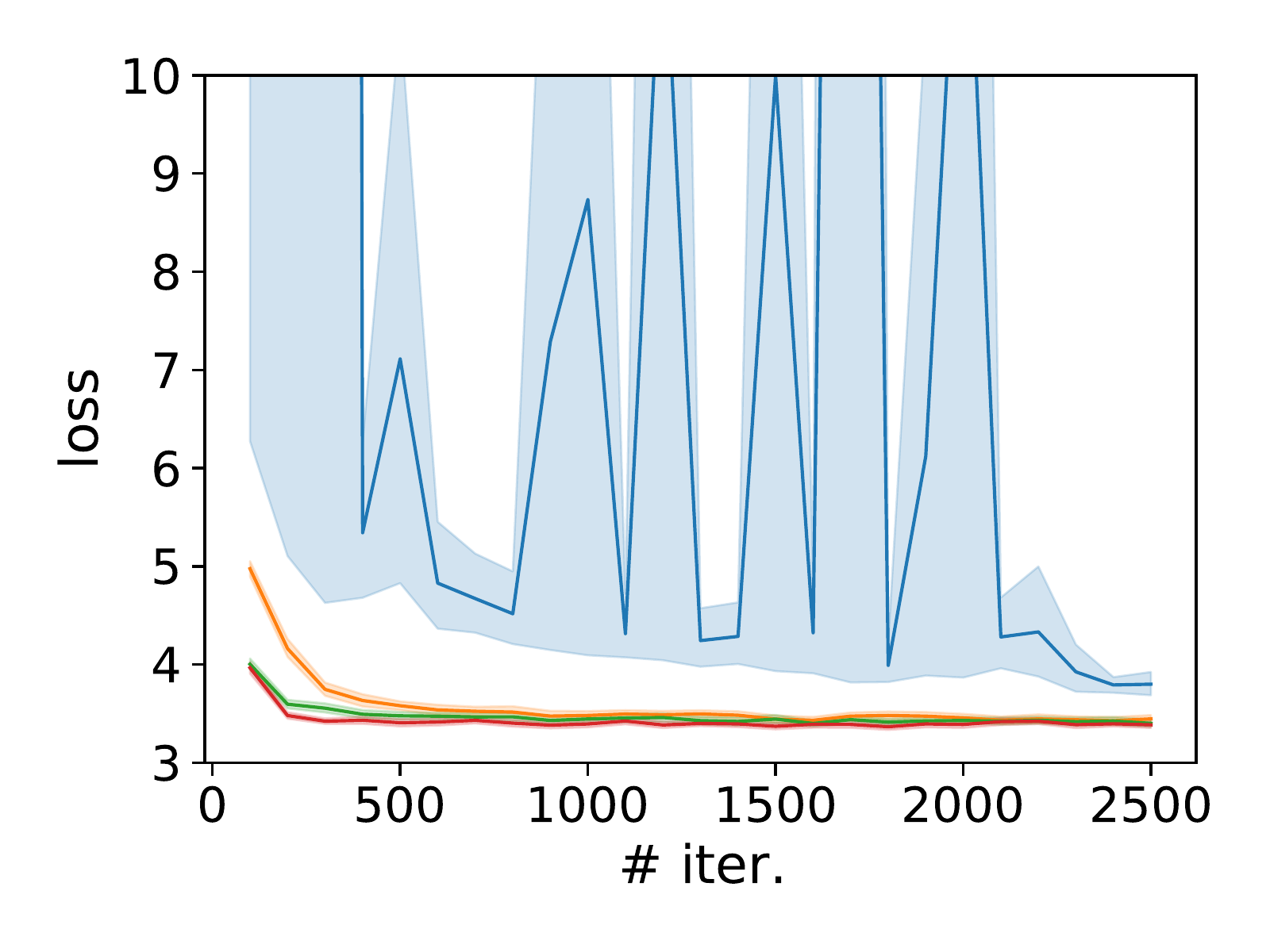}
       \caption{Training loss}
    \end{subfigure}
    \begin{subfigure}{0.23\textwidth}
       \includegraphics[width=\textwidth]{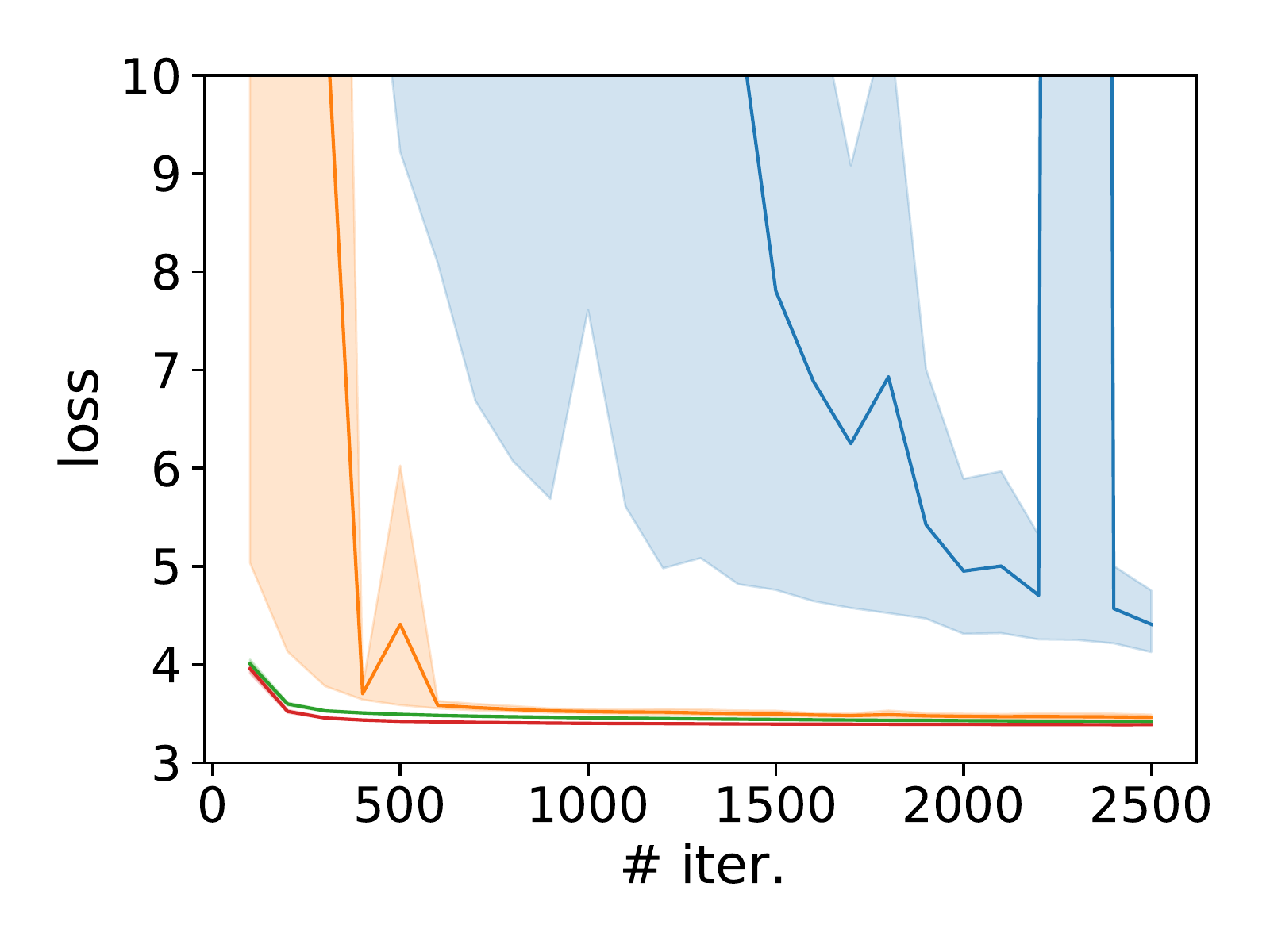}
       \caption{Test loss}
    \end{subfigure}
    \caption{Mean training and test loss over $100$ trails for NFs with different base distributions: \emph{normal} (blue), \emph{heavierTails} (orange), \emph{correctFamily} (green), and \emph{exactMarginals} (red). The shaded area depicts the $95\%$ confidence interval, which was computed using bootstrapping.
    We excluded \emph{normal} runs that achieved a final loss larger than $25$, which happened in $17$ out of $100$ runs.}
    \label{fig:trainingtest_performance}
\end{figure}

\paragraph{Training and test loss}
In Figure~\ref{fig:trainingtest_performance}, we plot the average training and test performance over $100$ trails. It is apparent that training using a base distribution with the correct type of tail behavior is beneficial.
First of all, we observe a significant gap between the test performance of the vanilla NF and the NFs with a heavy-tailed base distribution. 
Notice that in Figure~\ref{fig:trainingtest_performance} we excluded all runs with a final test loss of above $25$, which happened in 17 of the \emph{normal} runs and not once in the other cases. 
Furthermore, we clearly observe a much faster convergence and a more stable training procedure. The fluctuations and instabilities in the vanilla NF are due to tail-samples that have a massive effect on the likelihood in~\eqref{eq:mletraining}, which can be reduced by choosing base distributions with slower decaying tails \cite{alexanderson2020robust}.

\begin{figure}
    \centering
       \includegraphics[width=0.45\textwidth]{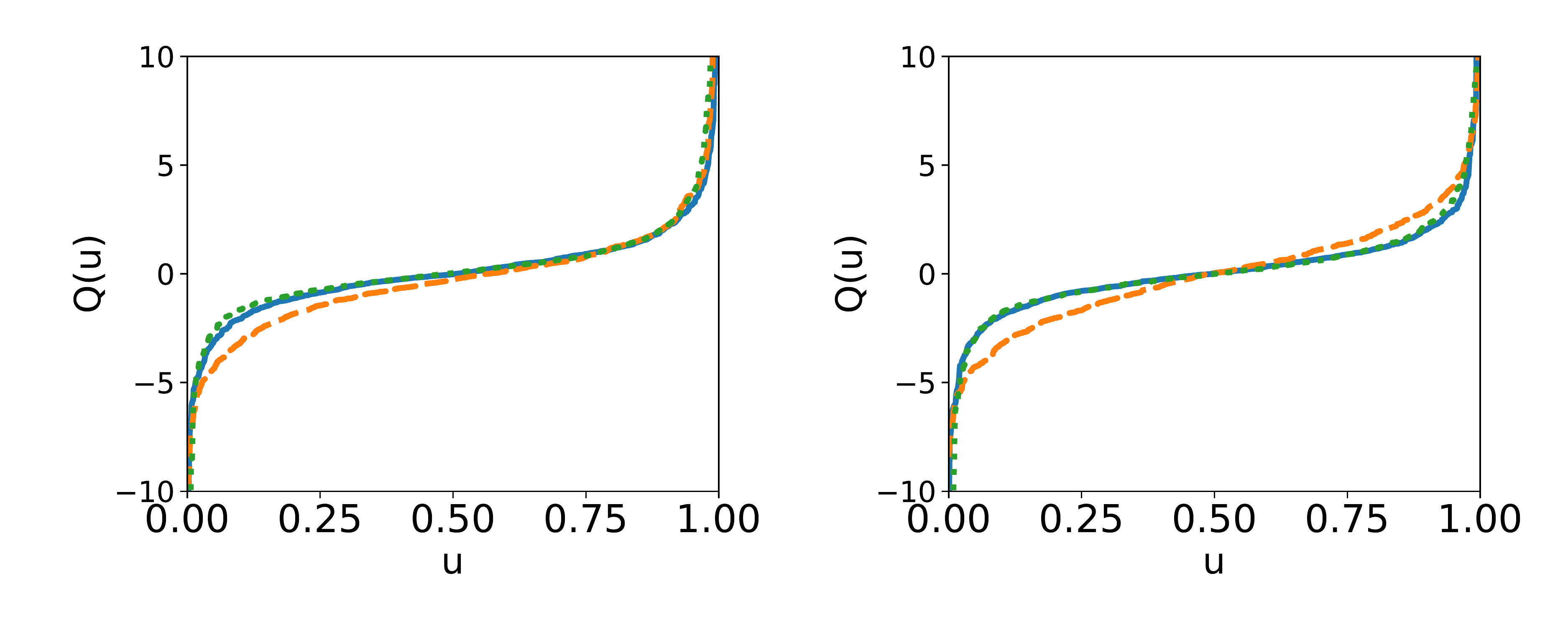}
    \caption{Estimated marginal quantiles in the case of \emph{normal} (orange, dashed) and \emph{exactMarginals} (green, dotted). The corresponding negative log-likelihoods are $4.00$ and $3.39$, respectively.}
    \label{fig:quantiles_marg}
\end{figure}

\paragraph{Learning the tails} 
To illustrate the ability to model the tails, we compared the estimated empirical quantile functions. We did so for both marginal distributions (Figure~\ref{fig:quantiles_marg}) but also for the distribution of $\Vert \rvx \Vert_2$ (Figure~\ref{fig:quantiles_norm} in the Appendix). 
In line with the findings by \citet{pmlr-v119-jaini20a}, we notice that the vanilla NF is not capable of modeling the quantiles of the target distribution. More precisely, we observe that the corresponding quantile function is steeper around its center and has shorter tails. This means that the distribution learned by the NF does not account for the heavy tails by directly modeling
them, but instead covers samples from the tails of the data distribution by being more widespread.
In contrast, the base distributions that took the tailedness of $\rvx$ into account, could achieve a much better fit to the quantiles, see Figure~\ref{fig:quantiles_marg_appendix} in the Appendix for further results. 

\begin{figure}
    \centering
    \begin{subfigure}{0.23\textwidth}
       \includegraphics[width=\textwidth]{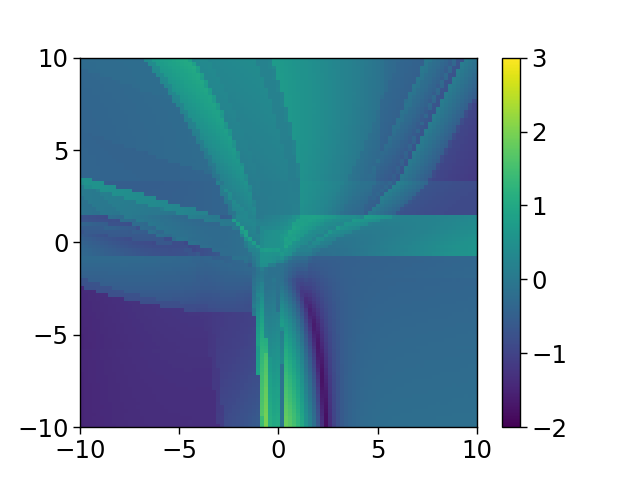}
       \caption{$T_\theta$ \emph{normal} ($3.54$)}
    \end{subfigure}
    \begin{subfigure}{0.23\textwidth}
       \includegraphics[width=\textwidth]{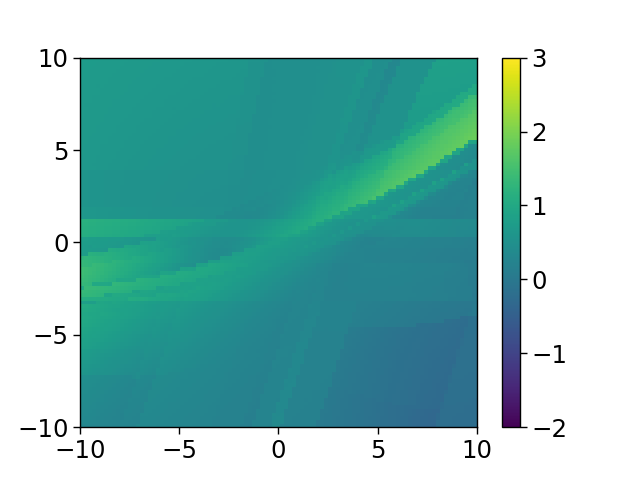}
   \caption{$T_\theta$ \emph{exactMarginals} ($3.39$)}
    \end{subfigure}
    \begin{subfigure}{0.23\textwidth}
       \includegraphics[width=\textwidth]{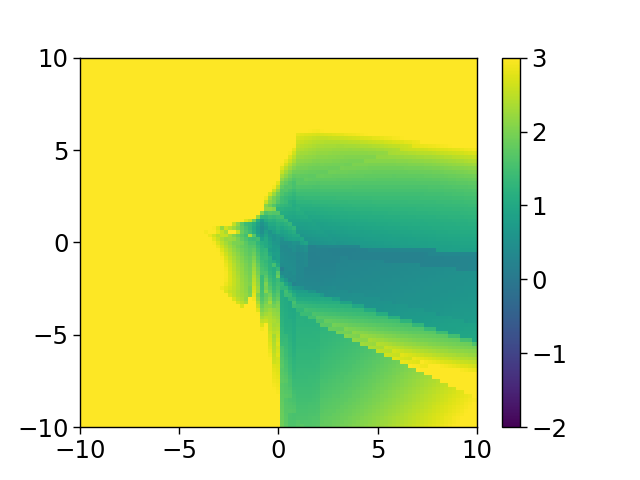}
       \caption{$T_\theta^{-1}$\emph{normal} ($3.54$)}
    \end{subfigure}
    \begin{subfigure}{0.23\textwidth}
       \includegraphics[width=\textwidth]{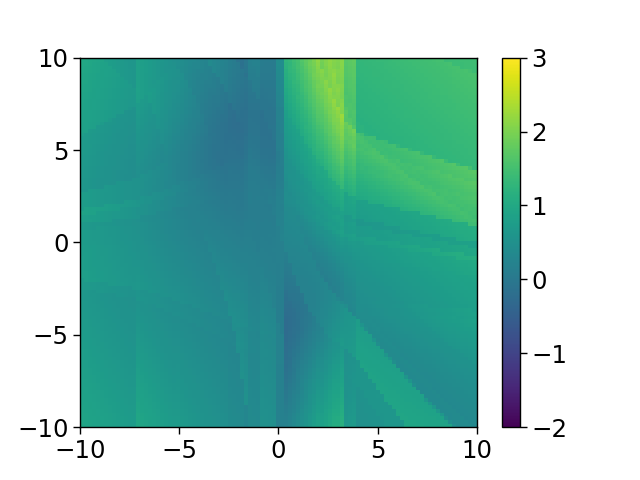}
       \caption{$T_\theta^{-1}$ \emph{exactMarginals} ($3.39$)}
    \end{subfigure}
    \caption{Examples for the Lipschitz surfaces of $T_\theta$ and $T_\theta^{-1}$ on a $\log$-scale. The corresponding negative log-likelihood is shown in brackets.}
    \label{fig:lipschitz_surface}
\end{figure}

\paragraph{Invertibility and numerical stability}
As investigated by \citet{pmlr-v130-behrmann21a}, the Bi-Lipschitz constant plays a fundamental role in the practical invertibility and numerical stability of NFs. 
To understand the learned transformation $T$ in terms of its Lipschitz continuity, we propose to study the \newterm{Lipschitz surface} of $T$.
Note that if $T$ is differentiable and $L$-Lipschitz, we can follow the derivation by \citet{pmlr-v130-behrmann21a} (in equation~(5) and (6) therein) to approximate 
\begin{align*}
    L &= \sup_{\vx \in \operatorname{supp}(\rvx)}\Vert J_T(\vx)\Vert_2 = \sup_{\vx \in \operatorname{supp}(\rvx)} \sup_{\Vert \vv \Vert_2=1}\Vert J_T(\vx)\vv \Vert_2 \\ &\approx \sup_{\vx \in \operatorname{supp}(\rvx)} \sup_{\Vert \vv \Vert_2=1} \frac{1}{\varepsilon} \Vert T(\vx) - T(\vx + \varepsilon \vv ) \Vert_2 \enspace, 
\end{align*}
where $\varepsilon$ is some small constant.
This motivates to consider an estimate\footnote{see Section~\ref{sec:comp_details} in the Appendix for details} of $\sup_{\Vert \vv \Vert_2=1}  \Vert T(\vx) - T(\vx + \varepsilon \vv ) \Vert_2/\varepsilon$ for $x \in \operatorname{supp}(\rvx)$ as a local surrogate for the Lipschitz-continuity of $T$.
Plotting these quantities for both, $T$ and $T^{-1}$, and $\vx \in [-10, 10]^2$ we obtain the Lipschitz surfaces, which are depicted in Figure~\ref{fig:lipschitz_surface}. We notice that the vanilla NF has many fluctuations in the local Lipschitz-continuity, while the proposed copula method leads to a well-behaved transformation. The inverse transformation $T_\theta^{-1}$ in the vanilla NF has exploding local Lipschitz constants, while---again---the proposed method results in a stable inverse transformation. 

\section{Discussion}
\label{sec:discussion}
In this work, we paved the way toward a general extension of  NF architectures using copulae. Synthetic experiments revealed that the modeling performance of NFs can be improved substantially by replacing the vanilla Gaussian base distribution by a base distribution that reflects basic properties of the data distribution more accurately. 
Of course, we have just scraped the surface of the underlying potential of the proposed approach: 
While we concentrate on the tail behavior of the marginals in this work, the general idea can potentially also be applied to incorporate other types of inductive bias, such as multimodality by choosing multimodal marginals, or symmetries and tail dependencies by selecting appropriate marginals and copulae.


Our experiments suggest that it is sufficient to have only a broad estimate of the marginals. 
As mentioned in Section~\ref{sec:framework}, one could also employ IFM to learn these before training the NF. 
However, the question of what is the best technique for choosing or estimating an appropriate marginal distributions and copula still requires further investigation. Nevertheless, we think that this flexibility brings additional improvement over the methods proposed by \citet{pmlr-v119-jaini20a} and \citet{alexanderson2020robust}. Of course, we have yet only gained preliminary results with our empirical study, which we plan to verify on real-world data and for different models in the future.

We believe that our analysis of the base functions can help to popularize NFs in a wide spectrum of domains.
One such application might be financial risk analysis, where it is essential to model tail dependencies. 

\section*{Acknowledgement}
This work was supported by the Deutsche Forschungsgemeinschaft (DFG, German Research Foundation) under Germany’s Excellence Strategy – EXC- 2092 CASA – 390781972. We also thank the anonymous reviewers for their careful reading and their useful
suggestions, which will help us extending this work to a full paper.
\bibliography{references.bib} 
\bibliographystyle{icml2021}

\clearpage
\appendix
\section{Examples of Copula Distributions}\label{sec:examples_copula}
\begin{figure}
    \centering
    \begin{subfigure}{0.35\textwidth}
       \includegraphics[width=\textwidth]{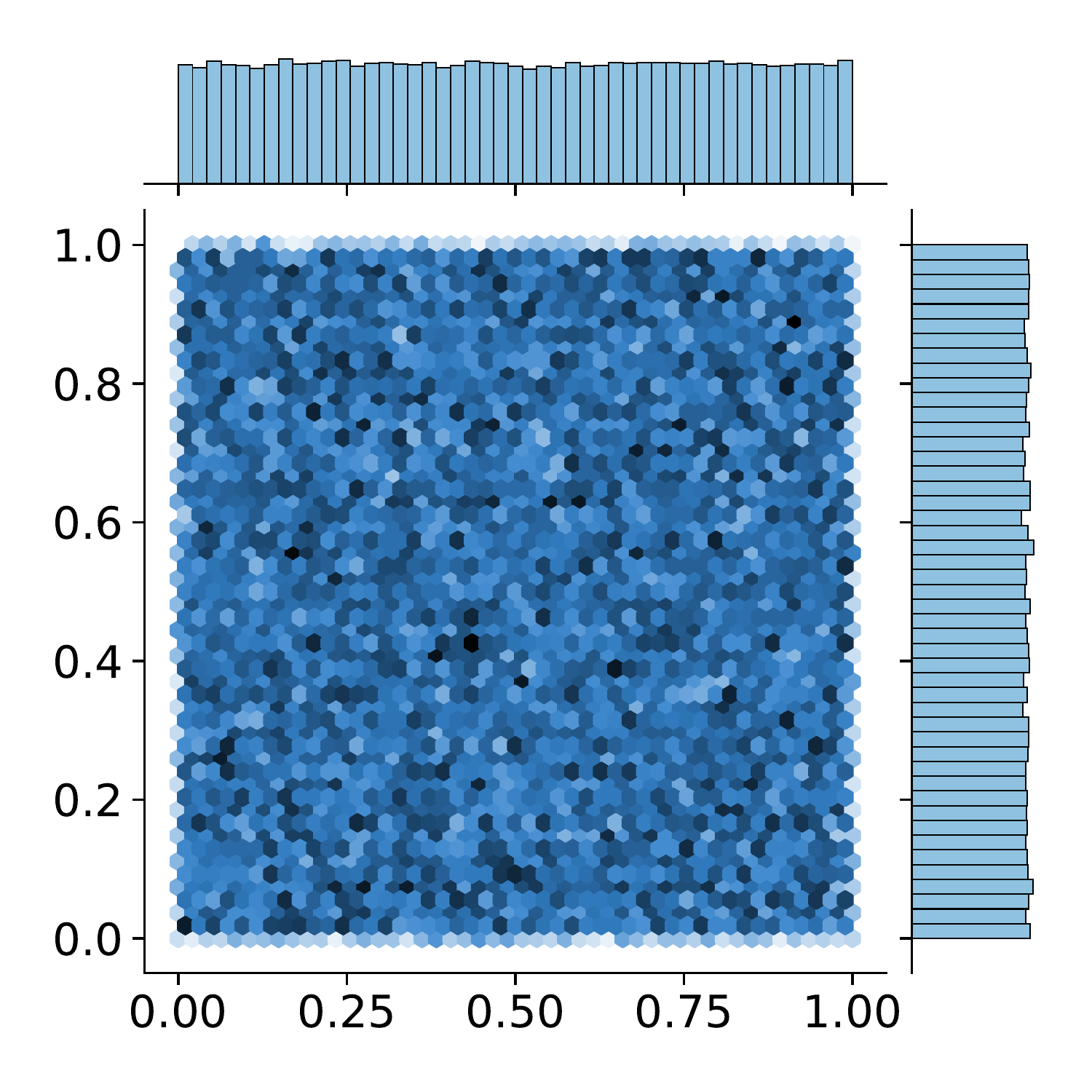}
       \caption{The Independence Copula}
    \end{subfigure}
    \begin{subfigure}{0.35\textwidth}
       \includegraphics[width=\textwidth]{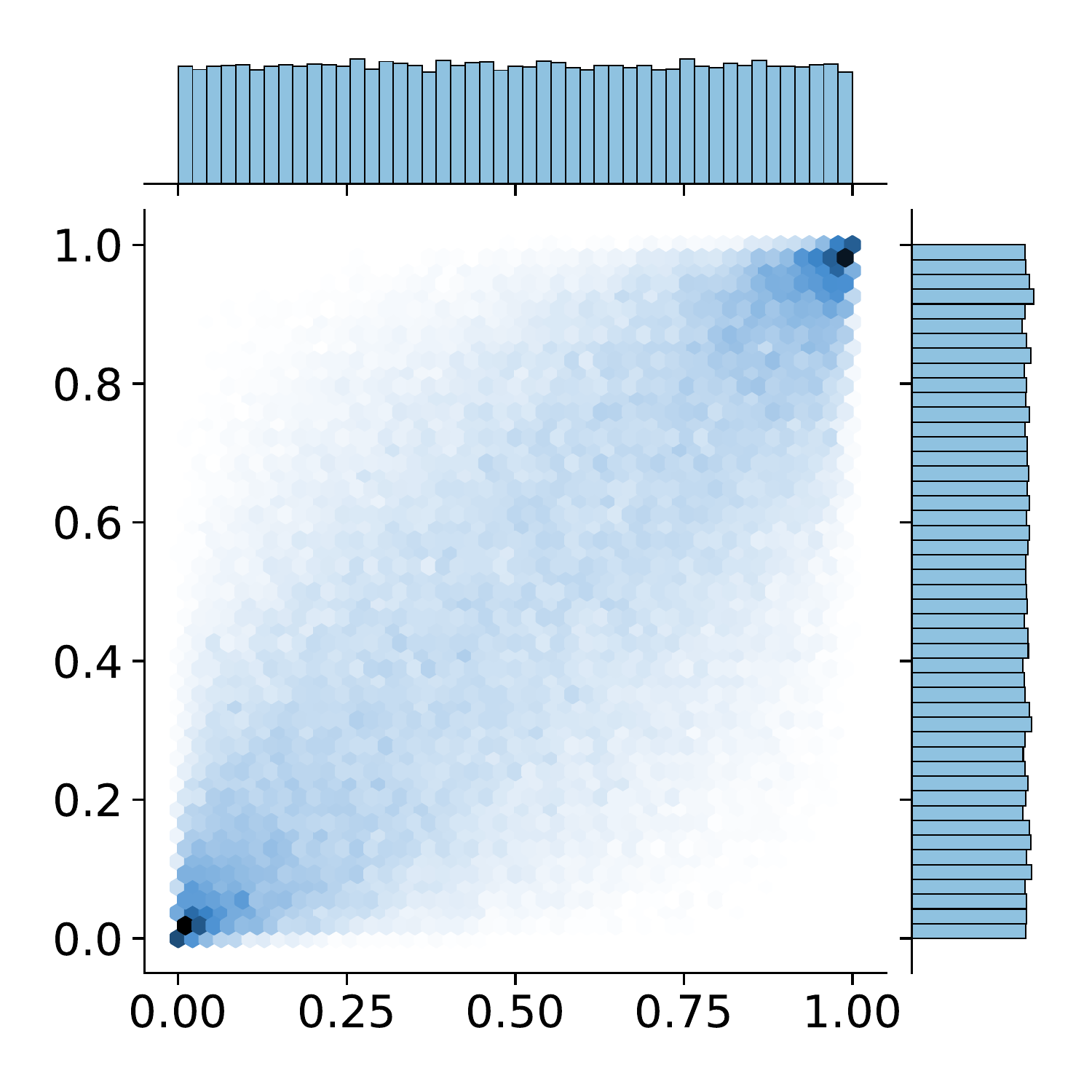}
       \caption{The Gaussian Copula with correlation $\rho_{12}=0.7$}
    \end{subfigure}
    \begin{subfigure}{0.35\textwidth}
       \includegraphics[width=\textwidth]{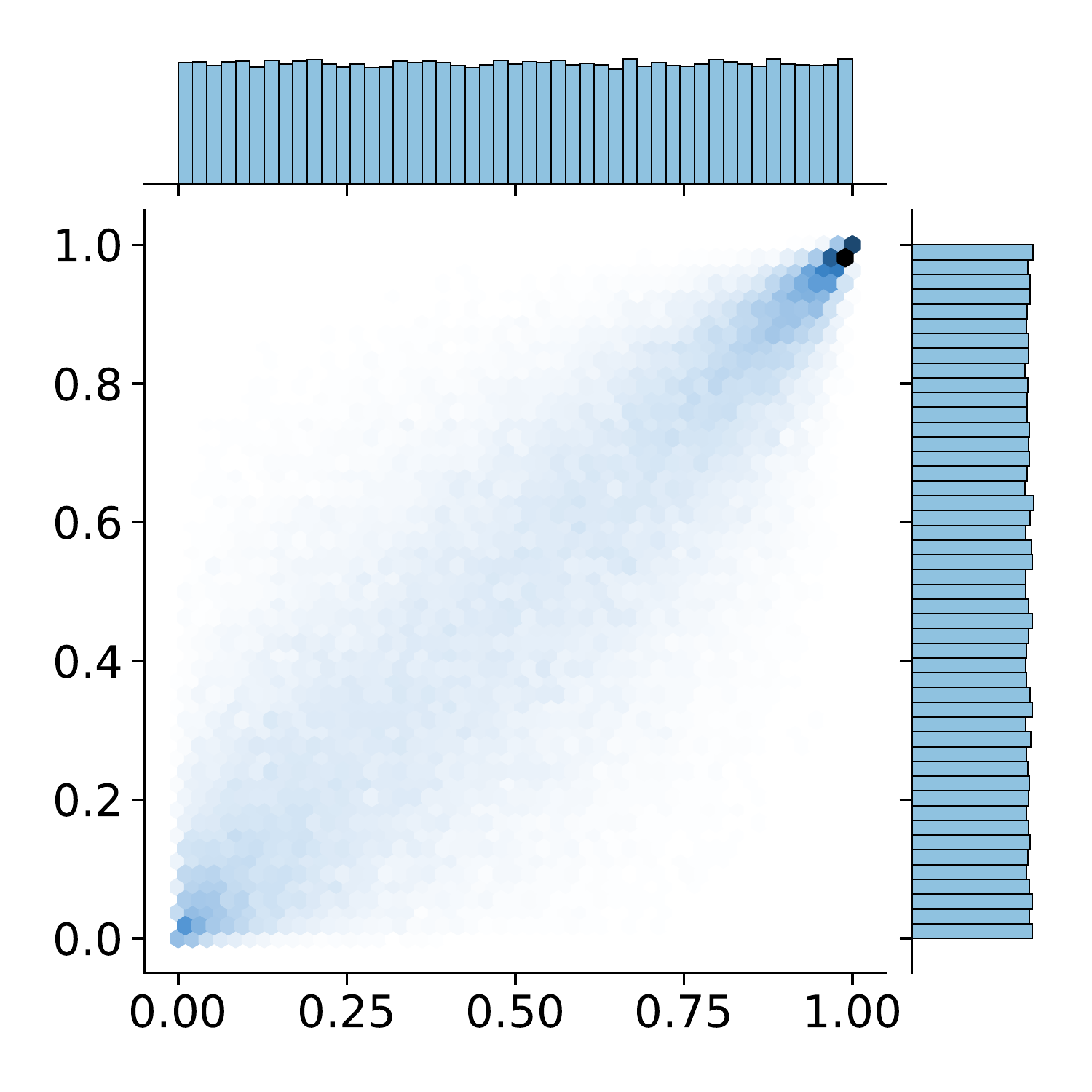}
       \caption{The Gumbel Copula with parameter $\rho=2.5$}
    \end{subfigure}
    \caption{Some popular Copulae.}
    \label{fig:copulae}
\end{figure}

\begin{figure}
	\centering
	\begin{subfigure}{0.35\textwidth}
	    \includegraphics[width=\textwidth]{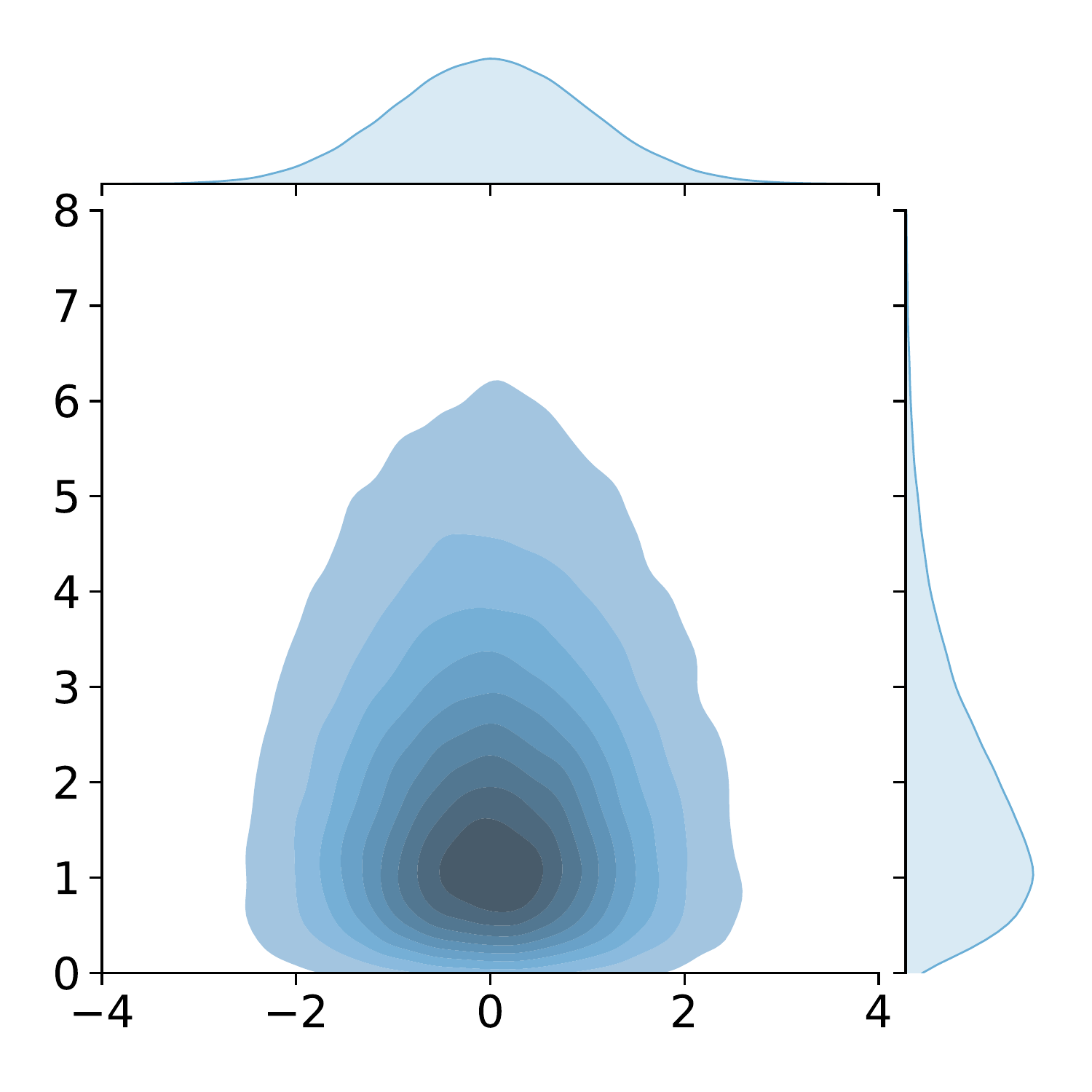}
	    \caption{The product distribution of a Gaussian and a Gamma marginal}
	\end{subfigure}
	\begin{subfigure}{0.35\textwidth}
	    \includegraphics[width=\textwidth]{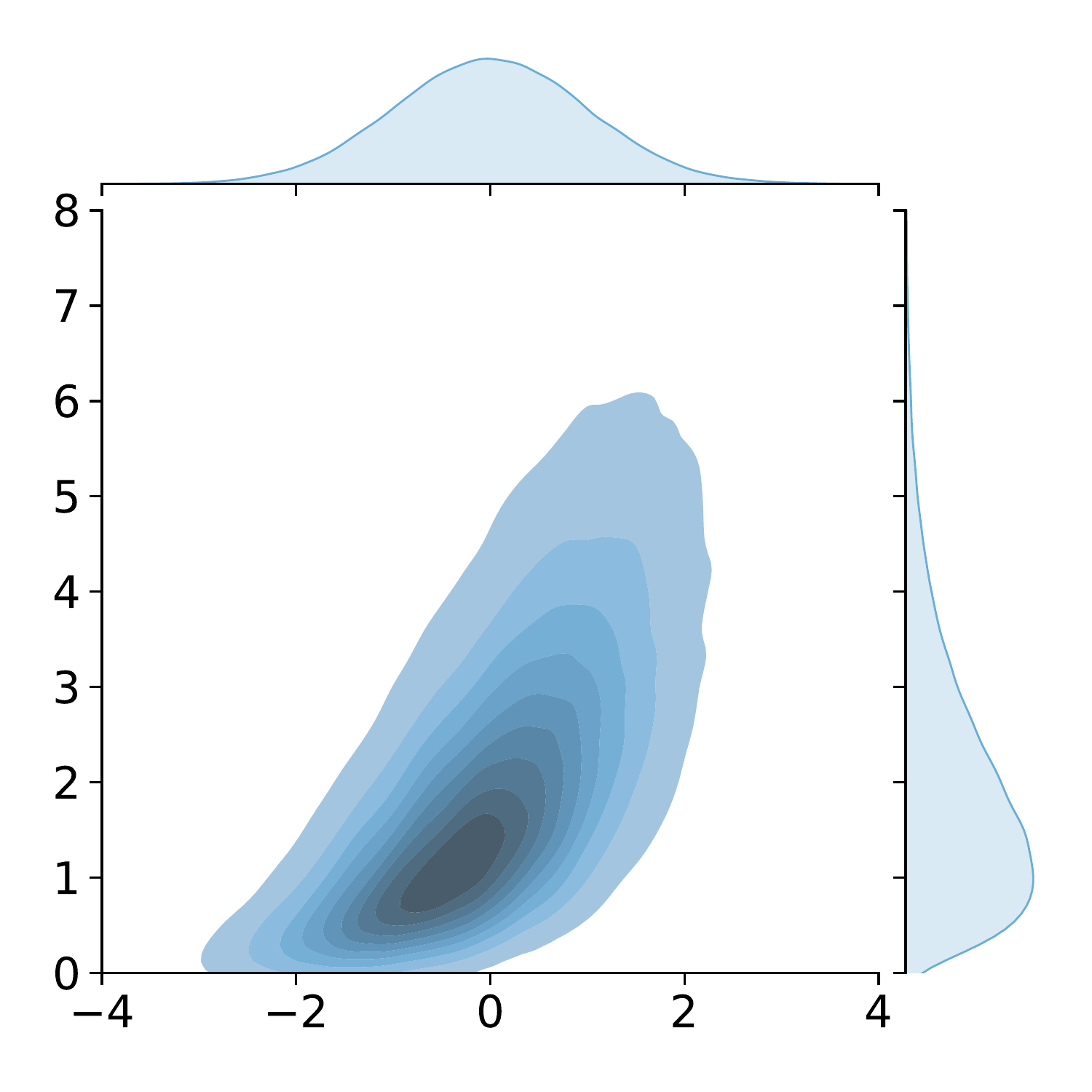}
	    \caption{The Gaussian Copula distribution with marginals from (a)}
	\end{subfigure}
	\begin{subfigure}{0.35\textwidth}
	    \includegraphics[width=\textwidth]{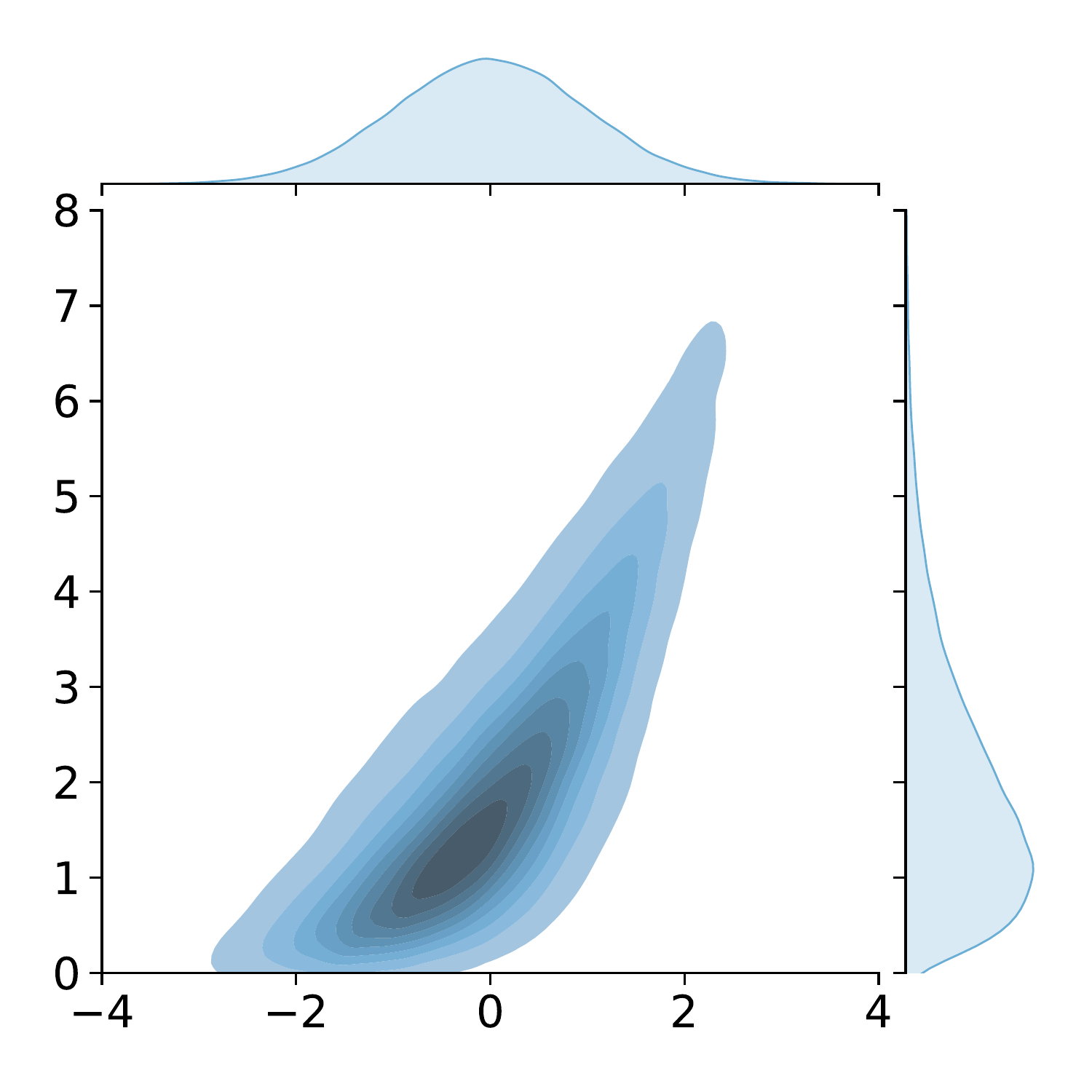}
	    \caption{The Gumbel Copula distribution with marginals from (a)}
	\end{subfigure}
	\caption{Some examples of distributions constructed via the copula approach.}
	\label{fig:example_copula}
\end{figure}

There is a wealth of different copula distributions, ranging from parametric distributions to semi-parametric and completely non-parametric copula models. In this Section, we give some insights into the construction of one class of copula distributions and illustrate some basic properties. 
\begin{exa}[Construction using Sklar's Theorem]\label{exa:copula_construction}
    There is a generic way to construct a copula according to Definition~\ref{defi:copula}. Consider any multivariate continuous and invertbile CDF $\Phi$ with marginal CDFs $\Phi_1,\dots, \Phi_D$. Then, following \eqref{eq:sklar_copula} in Sklar's Theorem~\ref{sklarstheorem}, we know that
    \begin{equation}
        \label{eq:generalCopula}
        C(\vu):= \Phi\bigl( 
            \Phi_1^{-1}(\vu_1), \dots ,\Phi_D^{-1}(\vu_D)
        \bigr)
    \end{equation}
    defines a valid copula. 
    Setting $\Phi$ to be the multivariate Gaussian distribution with correlation matrix $R$, we obtain the \newterm{Gaussian Copula}. If $R$ is the identity matrix we obtain the \newterm{Independence Copula}, which is simply the product of independent uniform distributions. Both copulae are visualized in Figure~\ref{fig:copulae}. Following the construction in \eqref{eq:generalCopula} and replacing the CDF, we can obtain other copulae,such as the t-Copula. Figure~\ref{fig:example_copula} visualizes one example for a distribution based on the Independence Copula and based on the Gaussian Copula.
\end{exa}

\begin{exa}[Copulae that induce Tail-Dependency]
    A crucial property of models in financial risk analysis is \newterm{tail dependency}. Roughly said, a tail dependency casts marginals to be dependent in a tail event. To give an example for a tail dependency, consider two \textit{essentially} independent marginal distributions, such as $\rvx_1=\{\text{debt status of your bank}\}$ and $\rvx_2=\{\text{trust for your bank}\}$. Usually, our trust for the bank is not essentially determined by the amount of debts it has. However, in a marginal tail event $\{\rvx=\text{bankrupt}\}$, of course, our trust drops rapidly. 
    Mathematically, the upper tail dependency, i.e. the dependency in upper-tail events, for a random variable $(\rvx_1, \rvx_2)$ is given by
    \begin{equation*}
        \lambda_U = \lim_{u\rightarrow 1^-}\mathbb{P}\bigl(\rvx_2> F_{\rvx_2}^{-1}(u)\, \vert \, \rvx_1 > F_{\rvx_1}^{-1}(u)\bigr)\enspace .  
    \end{equation*}
    Similarly, we can define the lower tail dependency $\lambda_L$. 
    One such copula that accounts for upper tail dependency is the \newterm{Gumbel Copula}, which is given by 
    \begin{equation}\label{eq:gumbel}
    C(\vu):= \exp\bigl( 
        -((-\log(\vu_1))^\rho + (-log(\vu_2))^\rho)^{1/\rho}
    \bigr) \enspace 
\end{equation}
    for $\rho \geq 1$. Figure~\ref{fig:copulae} shows a visualization of the Gumbel copula.
    One can show that $\lambda_U=2-2^{1/\theta}$ \citep{joe2014dependence}. Figure~\ref{fig:example_copula} visualizes the Gumbel copula distribution with Gaussian and Gamma marginals. We can observe a decent dependency---indicated by the peak pointing to the upper right---in the upper-tail events, which is mathematically described by $\lambda_U$.
\end{exa}
One could think of many more different properties that might be included using our copula approach, such as multi-modality and symmetry. 
However, it is still to be researched whether properties, such as the tail-dependency, are being preserved by the NF. Nonetheless, we think that fixing specific known properties in the base distribution facilitates training in acting as a type of regularization towards these given properties.

\section{Supplementary Experiments}\label{sec:suppexperiments}
In the following, we give some further empirical results that underpin our findings from the main paper.

\begin{figure}
    \centering
    \begin{subfigure}{0.23\textwidth}
        \includegraphics[width=\textwidth]{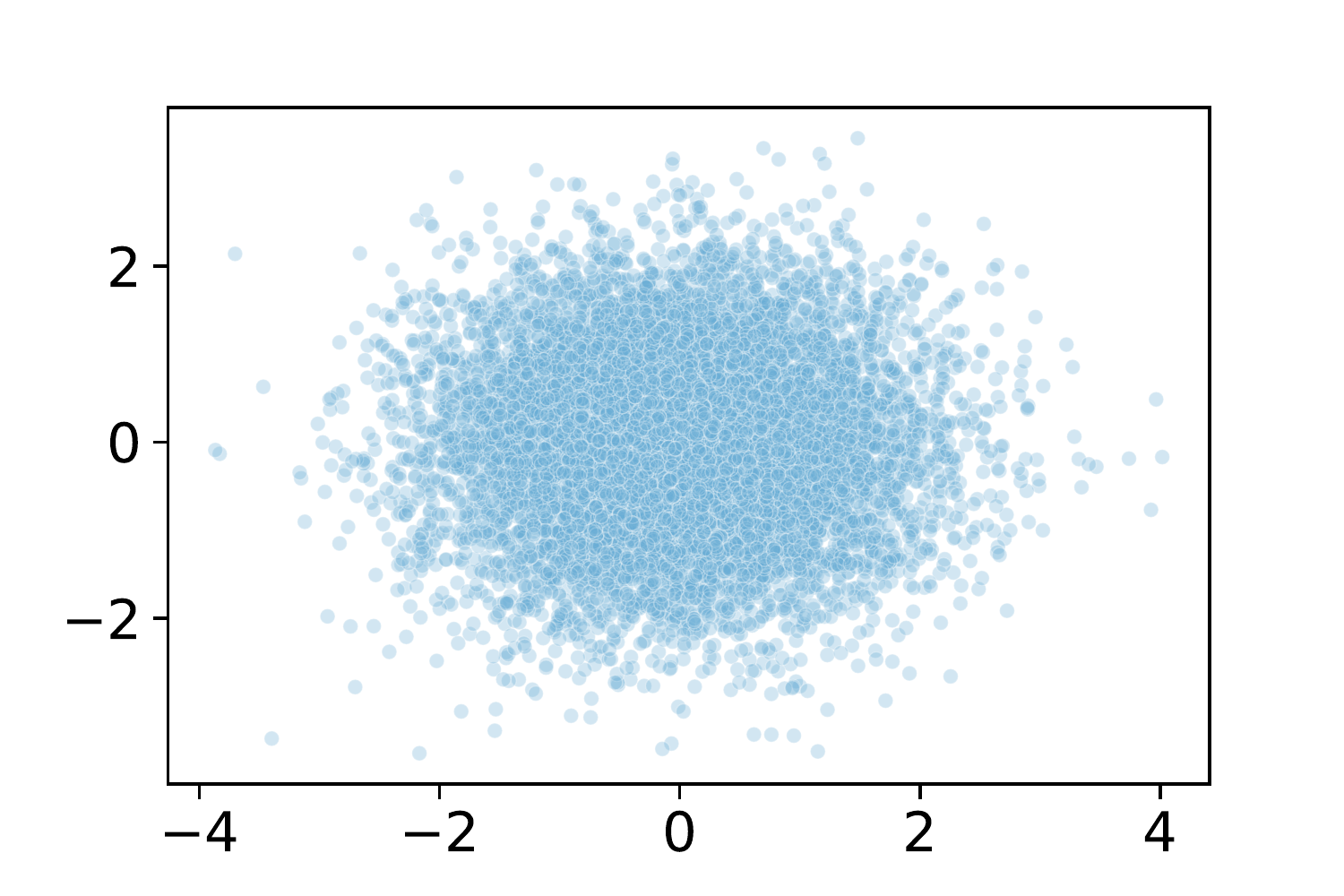}
        \caption{\emph{normal}}
        \label{fig:normal_dist}
    \end{subfigure}
    \begin{subfigure}{0.23\textwidth}
        \includegraphics[width=\textwidth]{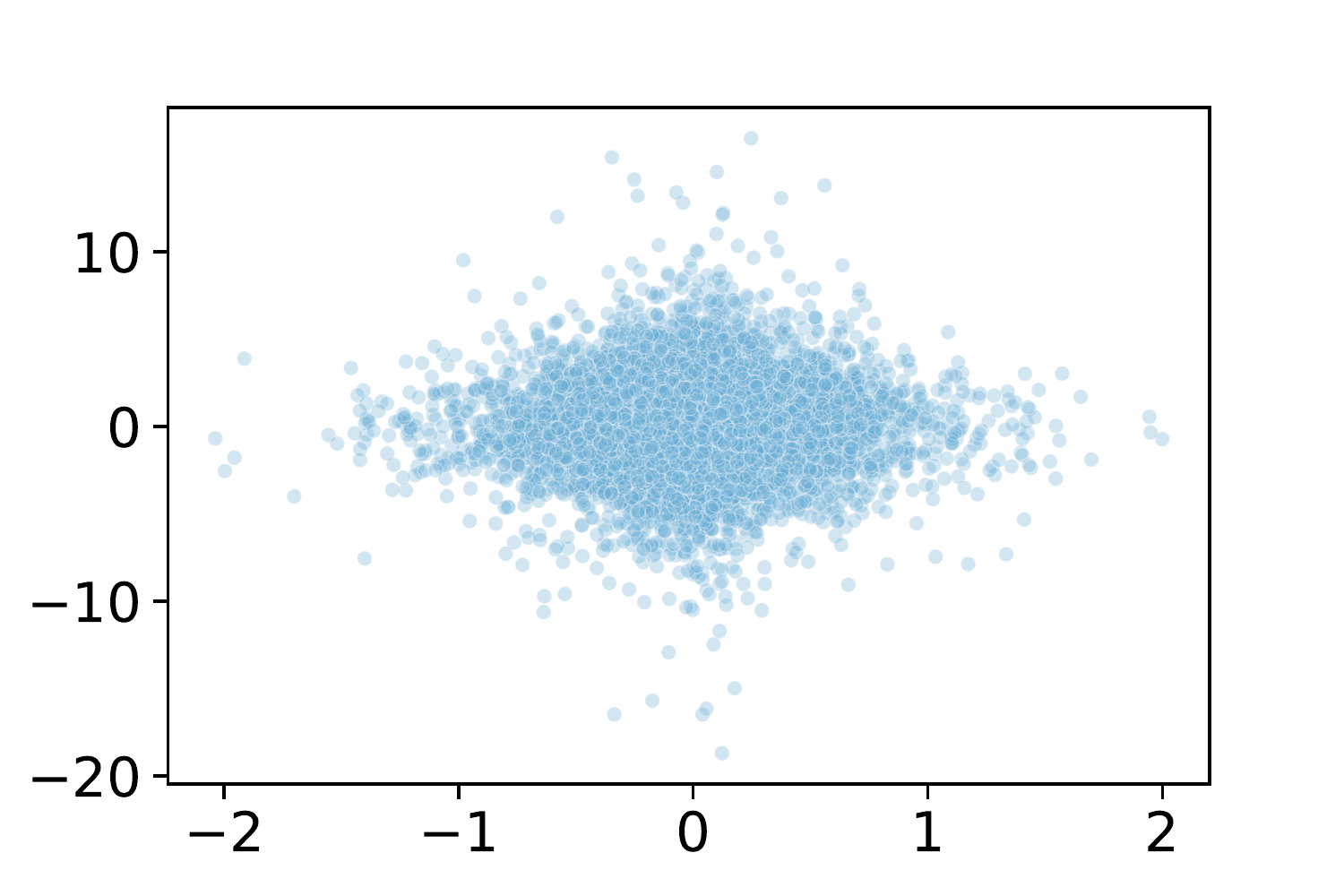}
        \caption{\emph{heavierTails}}
        \label{fig:laplaceT_dist}
    \end{subfigure} 
    \begin{subfigure}{0.23\textwidth}
        \includegraphics[width=\textwidth]{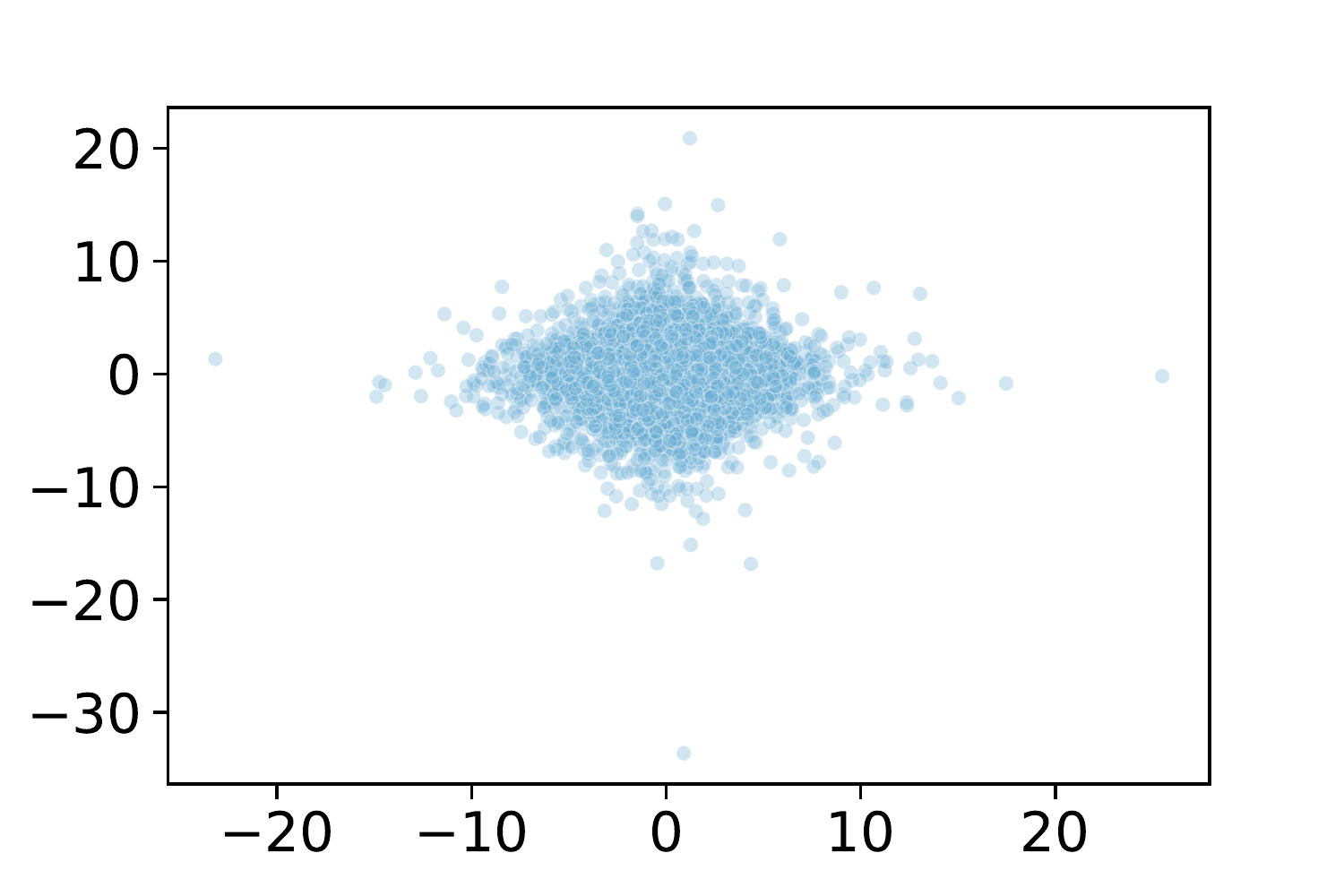}
        \caption{\emph{correctFamily}}
        \label{fig:TT_dist}
    \end{subfigure}
    \begin{subfigure}{0.23\textwidth}
        \includegraphics[width=\textwidth]{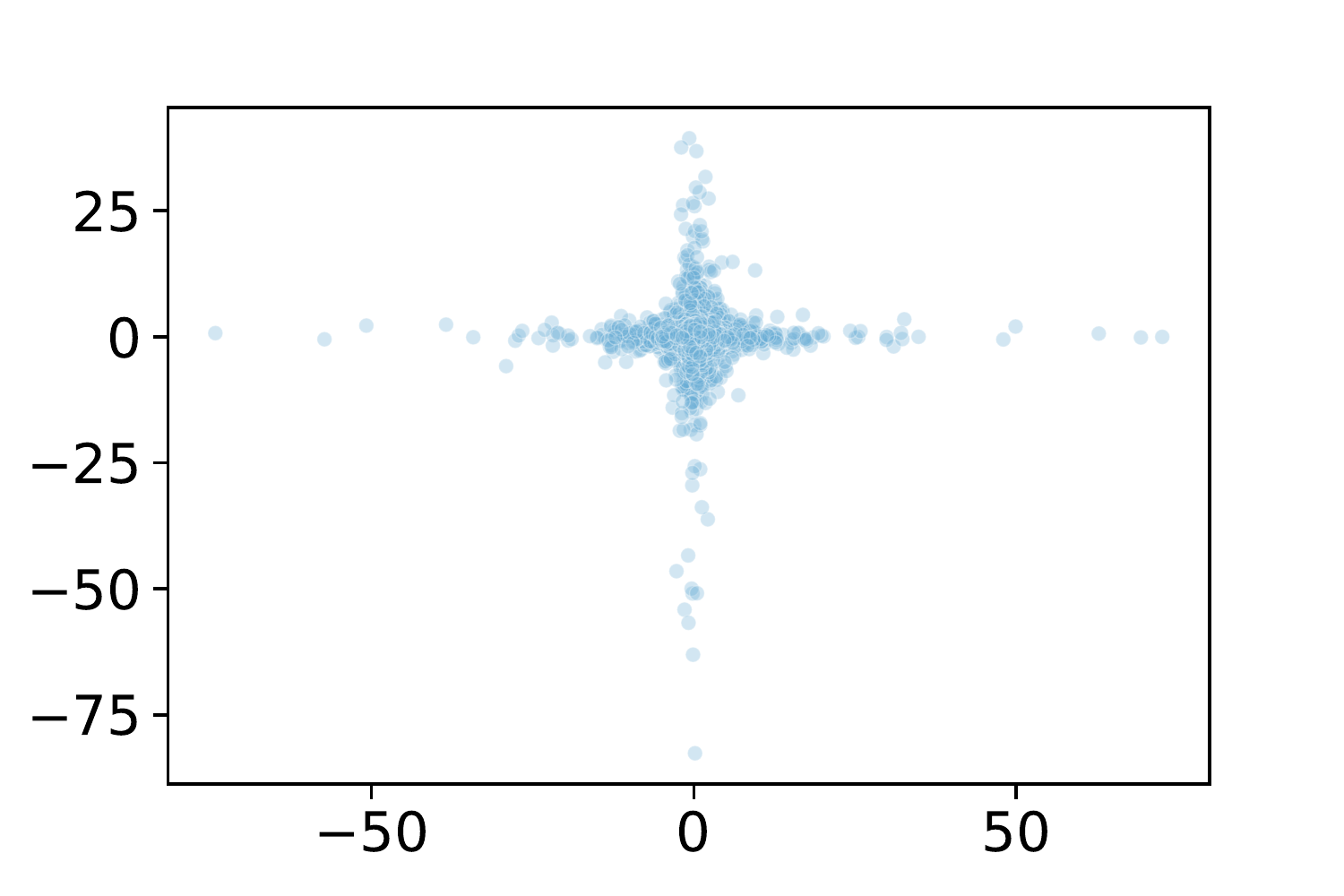}
        \caption{\emph{exactMarginals}}
        \label{fig:exact_marg}
    \end{subfigure}
    \caption{Samples from the 4 different base distributions that we used in our experiments.}
    \label{fig:base_dist}
\end{figure}

\begin{figure}
    \centering
    \begin{subfigure}[t]{0.23\textwidth}
        \includegraphics[width=\textwidth]{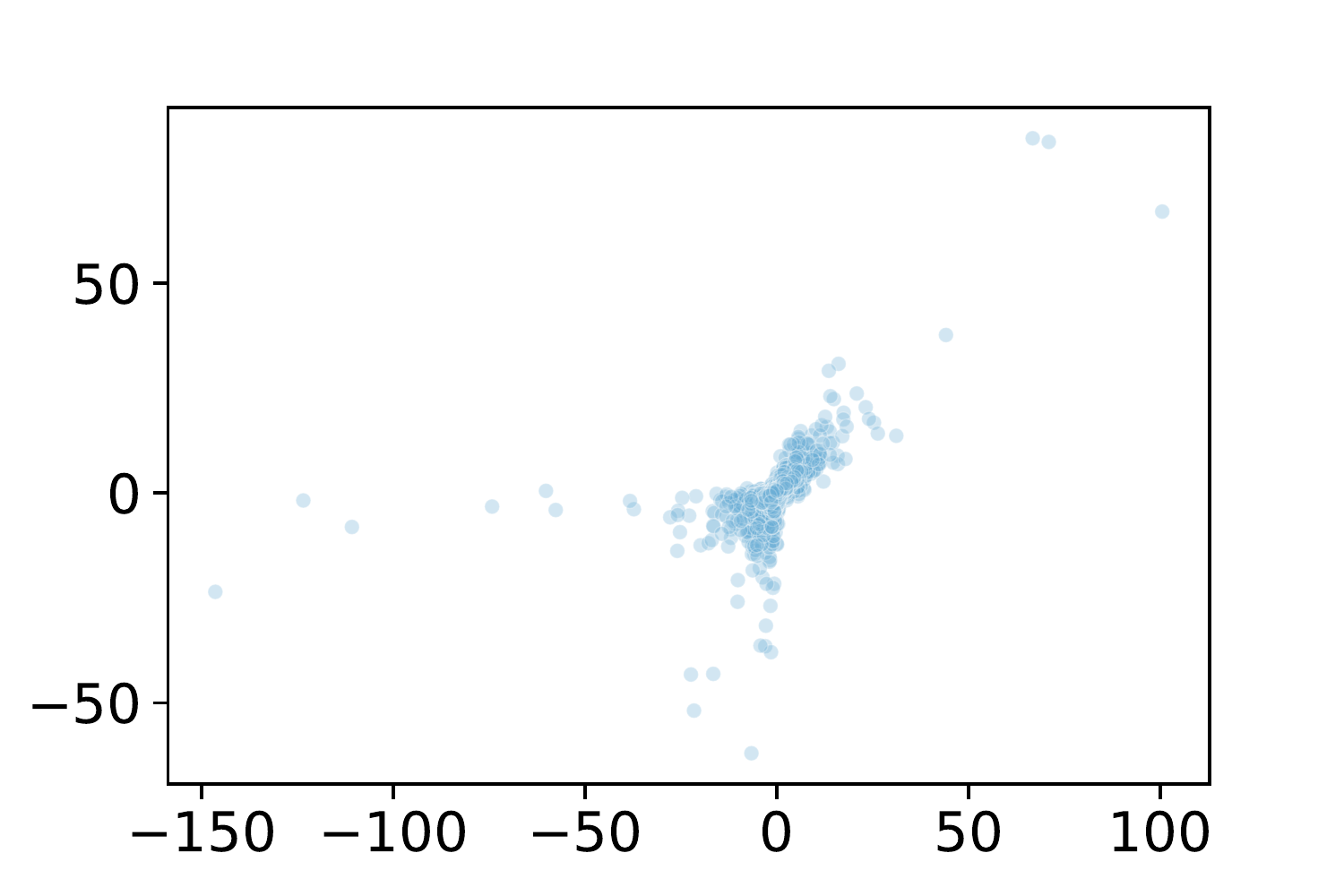}
        \caption{Gumbel Copula with $t_2 (0, 1)$-distributed marginals.}
        \label{fig:copula_dist}
    \end{subfigure}
    \begin{subfigure}[t]{0.23\textwidth}
       \includegraphics[width=\textwidth]{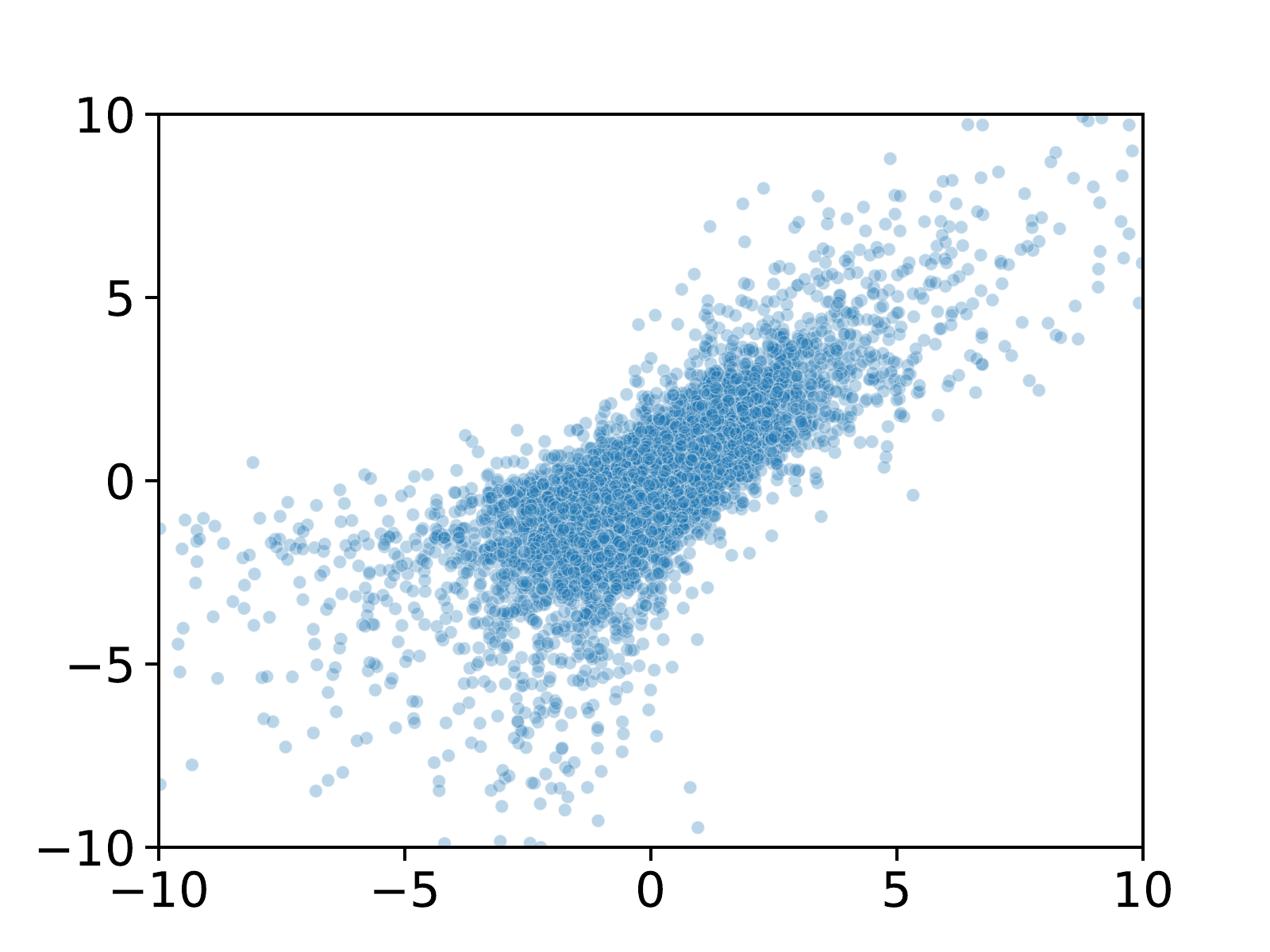}
        \caption{The distribution from~\ref{fig:copula_dist} zoomed in.}
    \end{subfigure}
    \caption{Samples from the target distribution.}
    \label{fig:target_dist}
\end{figure}

Figure~\ref{fig:base_dist} shows the investigated base distributions and Figure~\ref{fig:target_dist} visualizes the target distribution. 

\begin{figure}
    \centering
       \includegraphics[width=0.45\textwidth]{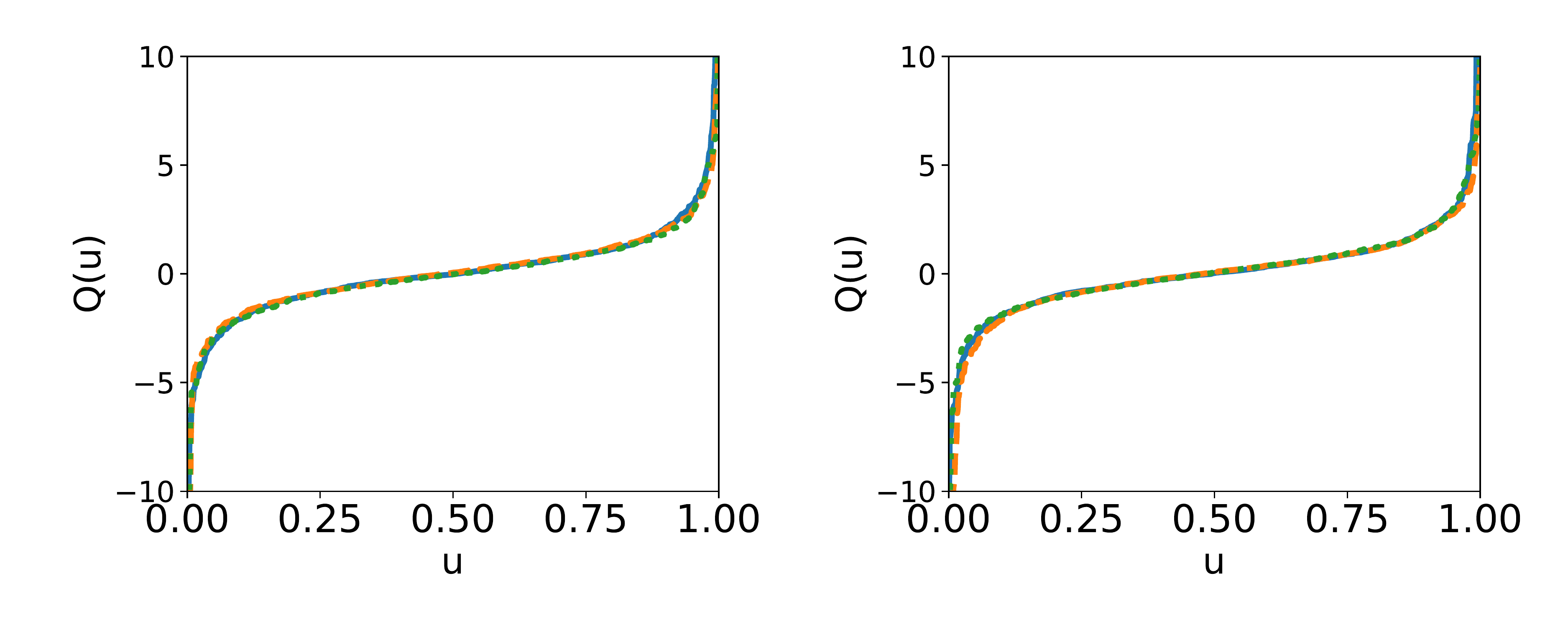}
    \caption{Estimated marginal CDFs in the case of \emph{heavierTails} (orange, dashed) and \emph{correctFamily} (green, dotted). The corresponding negative log-likelihoods are $3.44$ and $3.43$, respectively.}
    \label{fig:quantiles_marg_appendix}
\end{figure}

\begin{figure}
    \centering
    \begin{subfigure}{0.23\textwidth}
       \includegraphics[width=\textwidth]{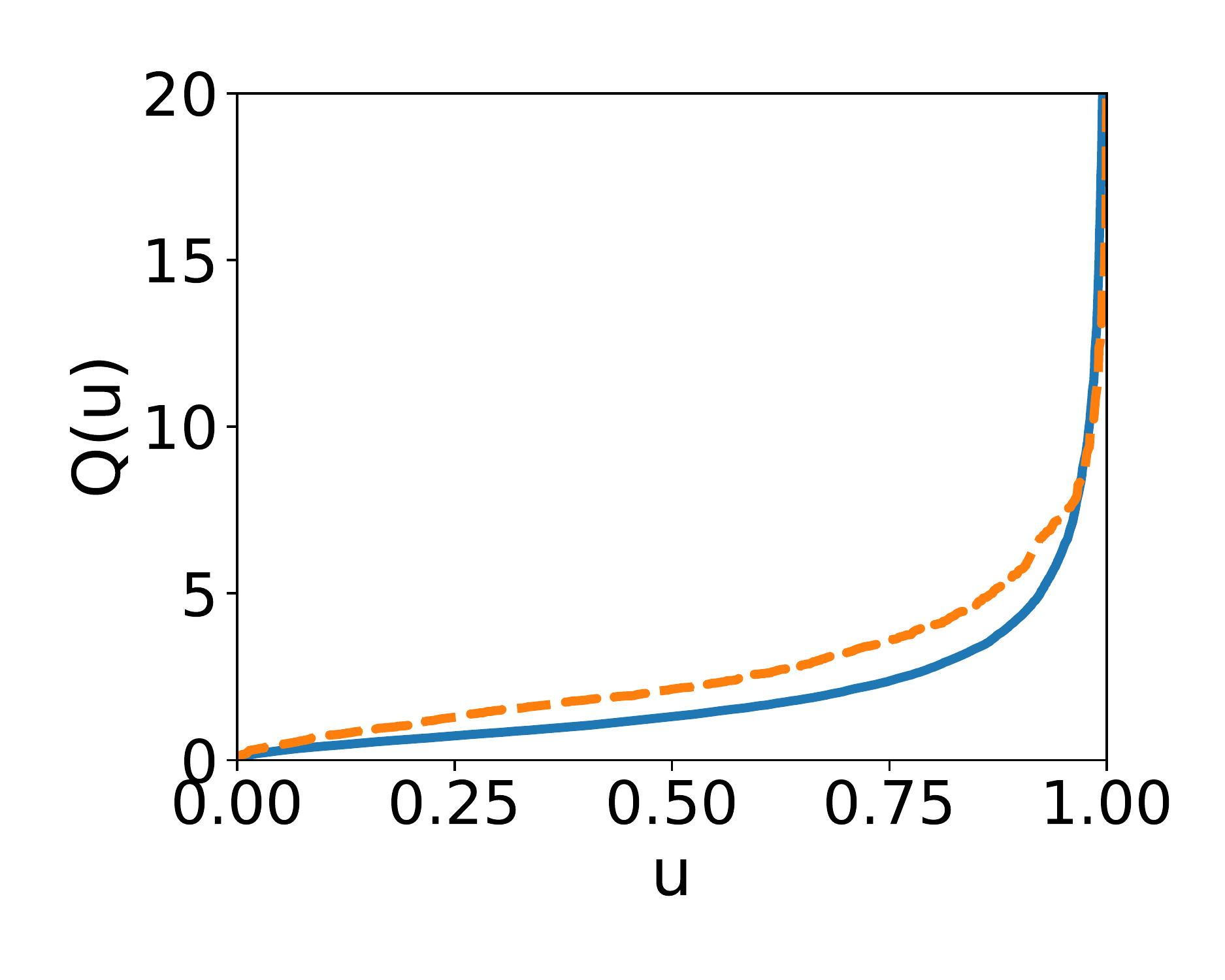}
       \caption{\emph{normal} ($4.00$)}
    \end{subfigure}
    \begin{subfigure}{0.23\textwidth}
       \includegraphics[width=\textwidth]{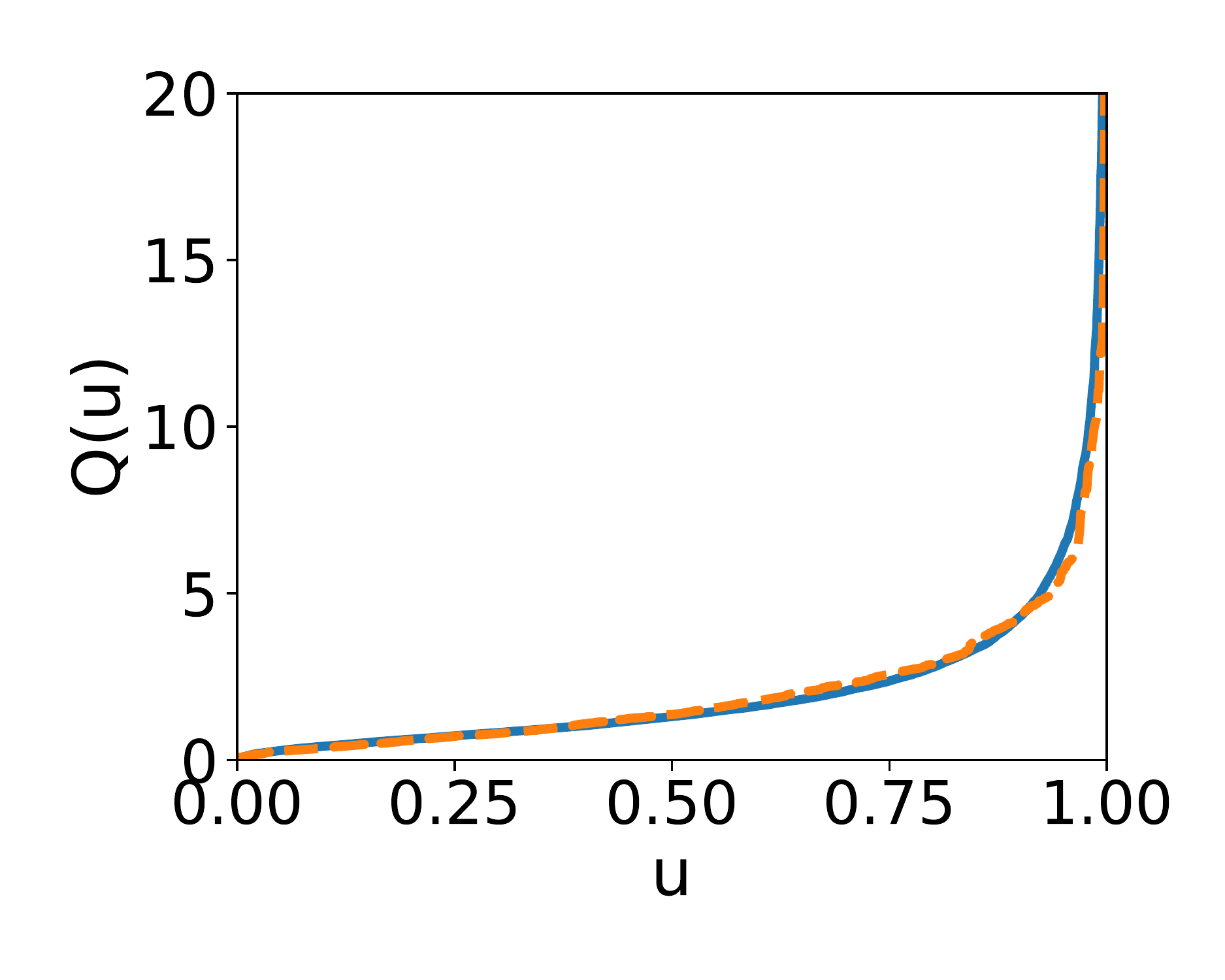}
       \caption{\emph{heavierTails} ($3.44$)}
    \end{subfigure}
    \begin{subfigure}{0.23\textwidth}
       \includegraphics[width=\textwidth]{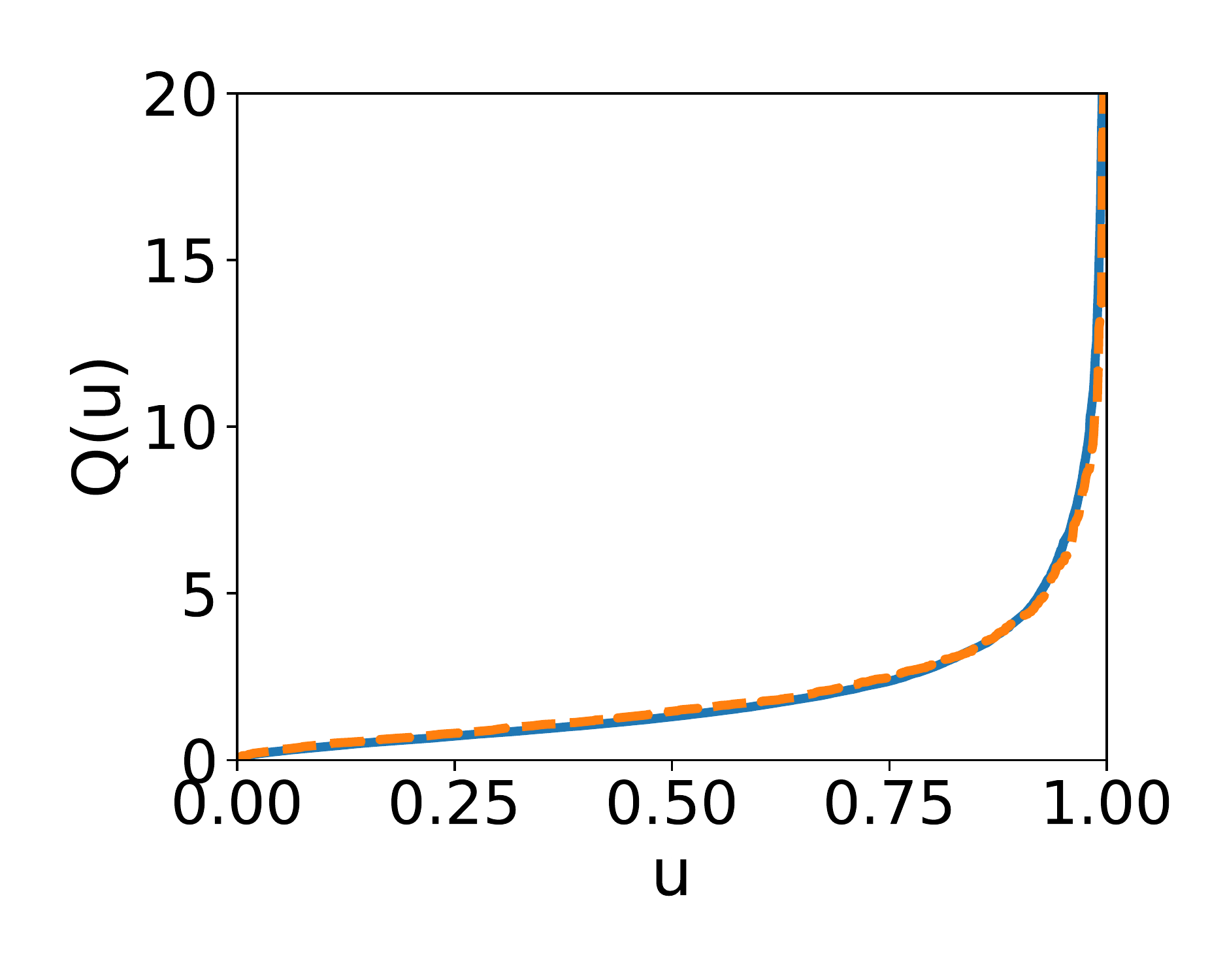}
       \caption{\emph{correctFamily} ($3.43$)}
    \end{subfigure}
    \begin{subfigure}{0.23\textwidth}
       \includegraphics[width=\textwidth]{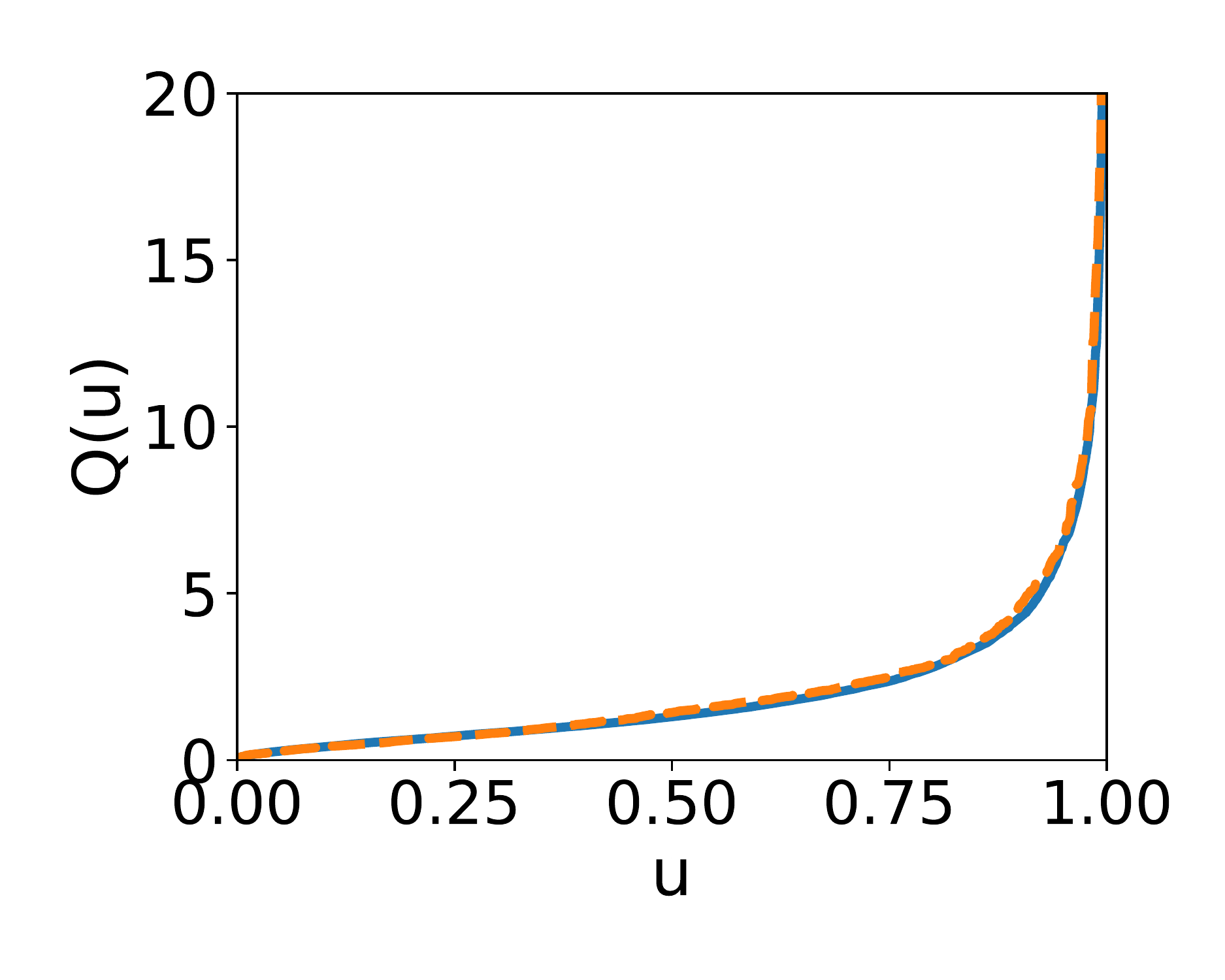}
       \caption{\emph{exactMarginals} ($3.39$)}
    \end{subfigure}
    \caption{Estimated quantiles of $\Vert \rvx \Vert_2$ using the different base distributions. The corresponding negative log-likelihood is shown in brackets.}
    \label{fig:quantiles_norm}
\end{figure}

Figure~\ref{fig:quantiles_marg_appendix} and Figure~\ref{fig:quantiles_norm} supplement the findings about the learned quantiles: Even the other heavy-tailed distributions---the cases \emph{heavierTails} and \emph{correctFamily}---are able to successfully approximate the quantiles. This observation suggests that it is sufficient to use a broad surrogate of the base distribution that captures the true tail behavior.

\begin{figure}
    \centering
    \begin{subfigure}{0.23\textwidth}
       \includegraphics[width=\textwidth]{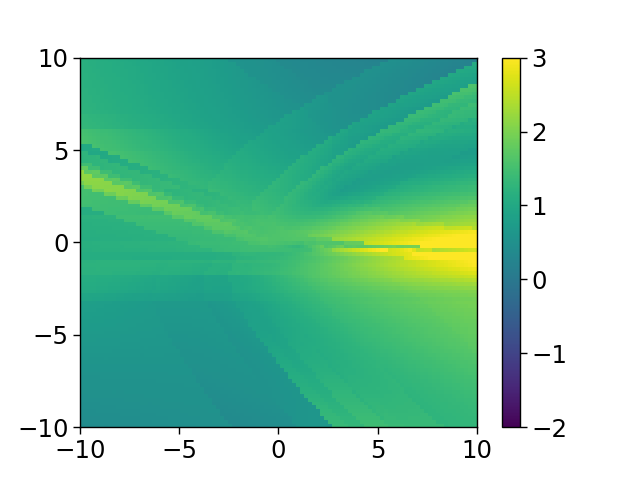}
       \caption{$T_\theta$ \emph{heavierTails} ($3.44$)}
    \end{subfigure}
    \begin{subfigure}{0.23\textwidth}
       \includegraphics[width=\textwidth]{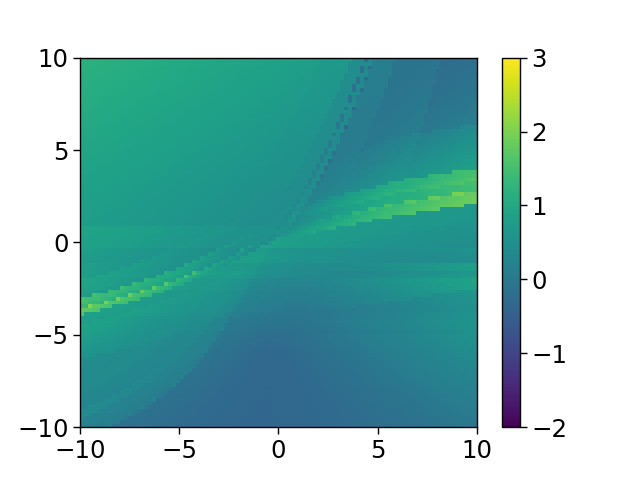}
   \caption{$T_\theta$ \emph{correctFamily} ($3.43$)}
    \end{subfigure}
    \begin{subfigure}{0.23\textwidth}
       \includegraphics[width=\textwidth]{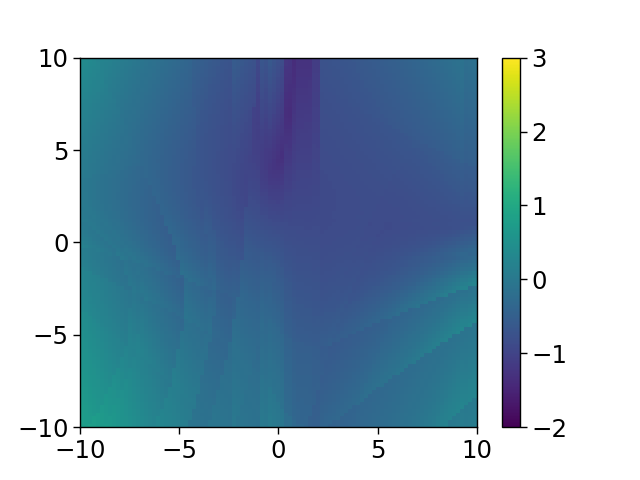}
       \caption{$T_\theta^{-1}$\emph{heavierTails} ($3.44$)}
    \end{subfigure}
    \begin{subfigure}{0.23\textwidth}
       \includegraphics[width=\textwidth]{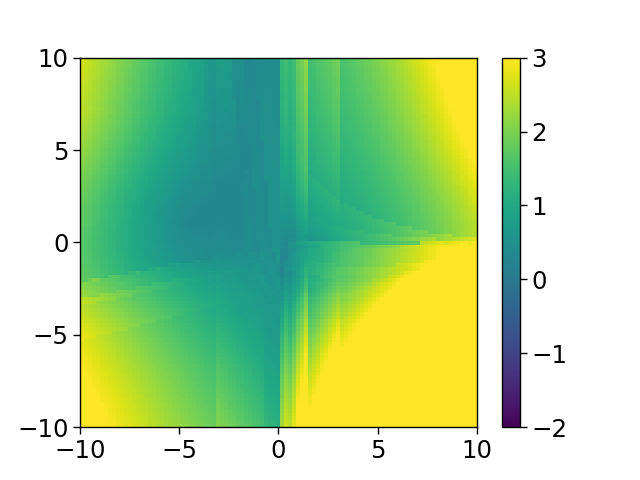}
       \caption{$T_\theta^{-1}$ \emph{correctFamily} ($3.43$)}
    \end{subfigure}
    \caption{Examples of the Lipschitz surfaces of $T_\theta$ and $T_\theta^{-1}$ on a $\log$-scale. The corresponding negative log-likelihood is shown in brackets.}
    \label{fig:lipschitz_surface_appendix}
\end{figure}

To address the invertibility and numerical stability of the learned transformation, we investigate the local Lipschitz constant as derived in Section~\ref{sec:experiments} in the main text. Figure~\ref{fig:lipschitz_surface_appendix} shows the supplementary Lipschitz surfaces of the cases \emph{heavierTails} and \emph{correctFamily}, which indicate a more regular transformation than the \emph{normal} case. Still, in the case \emph{correctFamily} we find large local Lipschitz constants in $T_\theta^{-1}$, which, however, are structured in accordance to the base distribution. In the vanilla NF this irregularity is less structured and, since these irregularities occur in high-probability areas of the base distribution, more relevant.

These results support our claim that choosing appropriately tailed base distribution can help in learning a numerically robust transformation.

\section{Computational Details}
\label{sec:comp_details}
In all of our experiments, we employed \newterm{Masked Autoregressive Flows} \citep{papamakarios2017masked} with 3 layers. Each layer contains a reverse permutation, followed by an autoregressive transformation with 4 hidden features. The code is based on the \emph{nflows} package \citep{nflows}. 

We trained the NFs on a sample of size $10\,000$. Optimization was carried out using the Adam optimizer with the \emph{PyTorch} default settings and a batch size of $128$. Test losses are evaluated based on $10\,000$ test samples.

The reported training and test losses (Figure~\ref{fig:trainingtest_performance}) have been averaged over $100$ runs and the depicted confidence intervals correspond to a confidence of $95\%$.

To estimate the Lipschitz surface we rely on an estimation of 
\begin{equation}
    \sup_{\Vert \vv \Vert_2=1} \frac{1}{\varepsilon} \Vert T(\vx) - T(\vx + \varepsilon \vv ) \Vert_2\enspace , \label{eq:sup_lipschitz}
\end{equation}
for some small constant $\varepsilon$. We chose $\varepsilon:=10^{-3}$. Further, we approximate \eqref{eq:sup_lipschitz} by
\begin{equation*}
    \max_{j\in \{1,\dots ,100\}} \frac{1}{\varepsilon} \Bigl\Vert T(\vx) - T(\vx - \varepsilon \frac{\vv^{(j)}}{\Vert \vv^{(j)}\Vert_2} \Bigr\Vert_2,
\end{equation*}
where $\vv^{(1)}, \dots \vv^{(100)}$ are i.i.d. samples from $\mathcal{N}(0, I)$, see \citet{pmlr-v130-behrmann21a} for further details. 

All code is provided with the submission and can further be accesses through \url{https://github.com/MikeLasz/Copula-Based-Normalizing-Flows}.
\end{document}